\documentclass{article}
\usepackage{array}
\newcolumntype{L}{>{\raggedright\arraybackslash}X}

\usepackage{iclr2027_conference,times}
\iclrfinalcopy   

\usepackage[utf8]{inputenc}
\usepackage[T1]{fontenc}
\usepackage{hyperref}
\usepackage{url}
\usepackage{booktabs}
\usepackage{amsfonts}
\usepackage{amsmath}
\usepackage{amssymb}
\usepackage{nicefrac}
\usepackage{microtype}
\usepackage{xcolor}
\usepackage{graphicx}
\usepackage{float}
\usepackage{placeins}
\usepackage{afterpage}  
\usepackage[most]{tcolorbox}
\usepackage{tikz}
\usetikzlibrary{positioning, arrows.meta, shapes.geometric, calc}
\graphicspath{{fig/}}
\usepackage[textsize=footnotesize]{todonotes}
\usepackage{multirow}
\usepackage{array}

\usepackage{color-edits}
\addauthor{vc}{blue}
\usepackage{tabularx}
\usepackage{makecell}
\usepackage{booktabs}

\usepackage{pifont}
\definecolor{cgreen}{RGB}{34,139,34}
\definecolor{cred}{RGB}{200,30,30}
\definecolor{cyellow}{RGB}{224,168,0}
\newcommand{\capfull}{\textcolor{cgreen}{\ding{51}}}
\newcommand{\capno}{\textcolor{cred}{\ding{55}}}
\newsavebox{\cappartbox}
\savebox{\cappartbox}{\tikz[baseline=-0.55ex]{\draw[densely dashed, cyellow, line width=0.9pt] (0,0) circle[radius=0.5ex];}}
\newcommand{\cappart}{\usebox{\cappartbox}}

\title{Synthetic Hospital: An Open, Verifiable,\\ Physician-Validated Longitudinal\\ EHR Benchmark}

\author{\begin{minipage}{\dimexpr\textwidth-2\tabcolsep\relax}\centering
Christine Park, Valerie Chen, Tim Dettmers\\[2pt]
\mdseries Carnegie Mellon University
\end{minipage}}

\begin{document}
\maketitle
\lhead{Preprint}   
\begingroup\makeatletter
\renewcommand\thefootnote{}
\long\def\@makefntext#1{\noindent#1}
\footnotetext{Code and data: \url{https://github.com/sparkcpark/synthetic_hospital}}
\makeatother\endgroup

\begin{abstract}
Frontier language models are rarely used in clinical workflows because the realistic, longitudinal benchmarks needed to develop them are scarce. Real electronic health record (EHR) data cannot be openly shared due to privacy, ethics or data use issues and it does not contain verifiable ground truth since the chart records only reflect what clinicians documented. We introduce Synthetic Hospital, an open, fully synthetic, fact-grounded longitudinal EHR benchmark that resolves the open sharing and verifiable ground truth barriers. Built entirely from public medical-education material with no protected health information, it comprises 1,268 longitudinal patients and 5,602 encounters, where every diagnosis, finding, and temporal relation is grounded in standard ontologies (ICD-10-CM, SNOMED CT, LOINC) and with a complete provenance chain back to its source medical education material. Synthetic Hospital is served through a simulated hospital record system that mirrors real EHR infrastructure (standard interoperability APIs, role-based access and function-calling interface). In a blinded review, physicians distinguished its records from real patient charts at near-chance rates (53\%). Across 10 frontier and open models, none approaches ceiling: the best model reconstructs a patient's longitudinal problem list with a severity-weighted F1 of 0.73, level with the mean of seven physicians on a matched subset but well below the best of them (0.89), and misses roughly half of clinically relevant findings when summarizing a chart. Overall, these results highlight that Synthetic Hospital is a difficult and realistic test of clinical AI performance. 
\end{abstract}

\section{Introduction}
\label{sec:intro}

Clinical AI has advanced rapidly on medical-knowledge benchmarks~\citep{singhal2023medpalm,nori2023medprompt,nori2024medprompto1}, yet those gains have not translated into routine use in clinical workflows~\citep{gong2025benchmarkreview}. A central reason is that the benchmarks used to develop and compare models do not reflect the settings in which clinical AI systems must operate. Real clinical work requires reasoning over patient records that span many encounters over months or years~\citep{fleming2024medalign,adams2024longhealth}, where electronic health records (EHR) are sparse and incomplete, may contain errors or inconsistencies~\citep{hripcsak2013phenotyping,weiskopf2013dataquality,bell2020noteerrors}, and clinically relevant evidence is distributed across notes, laboratory results, imaging reports, and other data stored in variable formats~\citep{rajkomar2018scalable,fleming2024medalign}. However, most widely used public medical benchmarks instead evaluate isolated multiple-choice questions, short-answer vignettes, or single-note tasks \citep{nori2024medprompto1,wang2025gpt5medical,singhal2023medpalm,nori2023medprompt,hendrycks2021mmlu}. 
While these benchmarks are useful for measuring medical knowledge, they are a poor match for the longitudinal systems clinicians require in practice. 

\afterpage{%
\begin{figure}[t!]
    \centering
    \includegraphics[width=\textwidth]{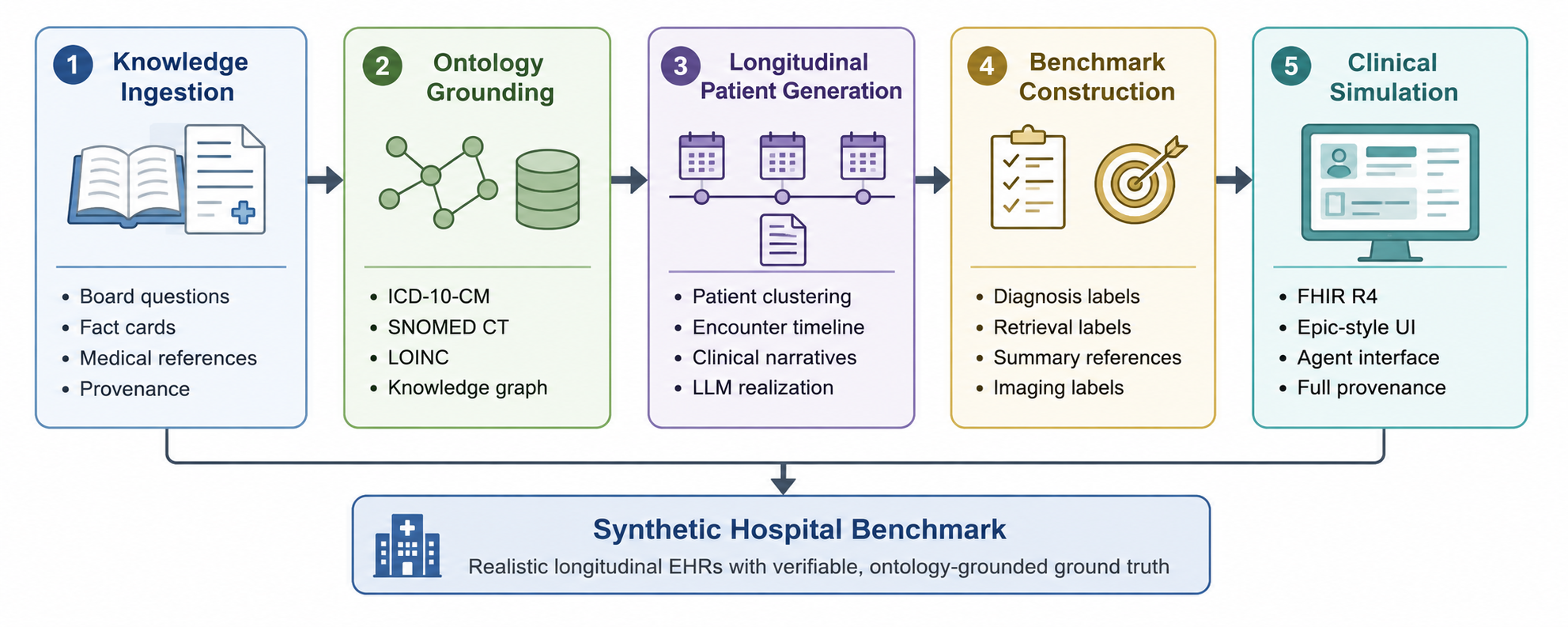}
    \caption{\textbf{Synthetic Hospital benchmark construction pipeline.} Public medical educational resources are transformed into a longitudinal, ontology-grounded benchmark through knowledge ingestion, graph construction, patient generation, ground-truth construction, and clinical simulation. Labels are derived before narrative generation, preserving provenance and independence from the rendering language model.}
    \label{fig:pipeline}
\end{figure}
}

However, using real EHR data as a starting point for a benchmark fails in two ways: it cannot be shared openly and it cannot consistently supply verifiable ground truth. On the first, privacy regulation, institutional review board (IRB) review, data use agreements (DUAs), de-identification requirements, and institution-specific governance all stand between a clinical dataset and an open benchmark, and even the de-identified EHR corpora that do exist are typically gated, narrow in scope, and difficult to redistribute \citep{johnson2023mimiciv,wornow2023ehrshot}. 

On the second, even with access to a real EHR, the chart reflects what was documented rather than the patient's complete clinical picture, including all relevant findings and their temporal relationships~\citep{hripcsak2013phenotyping}. Documentation errors~\citep{bell2020noteerrors}, missing notes~\citep{weiskopf2013dataquality}, and inconsistent coding~\citep{omalley2005icdaccuracy} make it difficult to tell whether a model reasoned incorrectly or whether the record itself was incomplete~\citep{alaa2025constructvalidity}, so the scoring that evidence retrieval, diagnosis, and longitudinal synthesis require has limited reliable referent.

In this paper, we introduce the \textbf{Synthetic Hospital}, an open, fully synthetic, knowledge-grounded longitudinal EHR benchmark that resolves both barriers. Although the patients are synthetic, their diagnoses, findings and clinical relationships are derived from medical educational sources and mapped to standard clinical ontologies, providing traceable provenance for the constructed patient state. Synthetic Hospital is built entirely from publicly available medical-education material containing no protected health information. Figure~\ref{fig:pipeline} summarizes the construction pipeline. 
Because every diagnosis, finding, and temporal relation is derived from source medical-education material, mapped to standard clinical ontologies (ICD-10-CM, SNOMED CT, and LOINC), and retained with its provenance (each encounter traces to a distinct source case, and no two patients share source material), the benchmark provides explicit ground truth for the constructed patient state and evaluation tasks. Of note, the record shown to a model still reads like a chart; it is the ground truth behind the record, not the record itself, that is complete. 

A key concern for synthetic clinical data is realism, where rule-based approaches to generating data may lack the narrative and longitudinal complexity needed to evaluate frontier models~\citep{walonoski2018synthea}. For an evaluation benchmark, our primary concern is whether individual records constitute plausible, clinically coherent test cases rather than whether the synthetic population reproduces real-world epidemiology. We evaluate this record-level realism directly: in blinded review, licensed physicians distinguished synthetic from real patient records only at near-chance rates ($53\%$; Section~\ref{sec:realism}). We separately characterize distributional properties in Appendix~\ref{app:casemix}, showing that Synthetic Hospital preserves the education-derived case mix of its source corpus and clinically expected comorbidity structure while, by design, differing from a population-calibrated cohort. 

\afterpage{%
\begin{table}[t!]
\centering
\small
\setlength{\tabcolsep}{4pt}
\resizebox{\textwidth}{!}{%
\begin{tabular}{@{}lcccccc@{}}
\toprule
\textbf{Benchmark} & \textbf{Open data} & \textbf{Longitudinal} & \textbf{Retrieval} & \textbf{Problem list} & \textbf{Summarization} & \textbf{EHR operation} \\
\midrule
\multicolumn{7}{@{}l}{\emph{Knowledge / QA}} \\
MedQA \citep{jin2021medqa} & \capfull & \capno & \capno & \capno & \capno & \capno \\
MedXpertQA \citep{zhou2025medxpertqa} & \capfull & \capno & \capno & \capno & \capno & \capno \\
HealthBench \citep{arora2025healthbench} & \capfull & \capno & \capno & \capno & \capno & \capno \\
MedHELM \citep{bedi2025medhelm} & \cappart & \capno & \capno & \capno & \cappart & \capno \\ 
\addlinespace
\multicolumn{7}{@{}l}{\emph{Clinical documentation / NLP}} \\
BRIDGE \citep{wu2025bridge} & \cappart & \capno & \capno & \capno & \capfull & \capno \\ %
ACI-Bench \citep{yim2023acibench} & \capfull & \capno & \capno & \capno & \cappart & \capno \\
ProbSum \citep{gao2023probsum} & \capno & \capno & \capno & \cappart & \cappart & \capno \\ 
\addlinespace
\multicolumn{7}{@{}l}{\emph{Long-document / vignette}} \\
LongHealth \citep{adams2024longhealth} & \capfull & \cappart & \cappart & \capno & \capno & \capno \\ 
\addlinespace
\multicolumn{7}{@{}l}{\emph{Agent / simulation}} \\
AgentClinic \citep{schmidgall2024agentclinic} & \capfull & \capno & \capno & \capno & \capno & \capno \\
SDBench \citep{nori2025sdbench} & \capno & \capno & \capno & \capno & \capno & \capno \\ 
MedAgentBench \citep{jiang2025medagentbench} & \capfull & \cappart & \cappart & \capno & \capno & \capfull \\ 
\addlinespace
\multicolumn{7}{@{}l}{\emph{Longitudinal real-EHR}} \\
MedAlign \citep{fleming2024medalign} & \capno & \capfull & \cappart & \capno & \capfull & \capno \\ 
ER-Reason \citep{mehandru2025erreason} & \capno & \capfull & \cappart & \capno & \cappart & \capno \\ 
EHRSQL \citep{lee2023ehrsql} & \capno & \capno & \cappart & \capno & \capno & \cappart \\
EHRSHOT \citep{wornow2023ehrshot} & \capno & \capfull & \capno & \capno & \capno & \capno \\
TIMER \citep{cui2025timer} & \capno & \capfull & \capno & \capno & \cappart & \capno \\
\midrule
\textbf{Synthetic Hospital (ours)} & \capfull & \capfull & \capfull & \capfull & \capfull & \capfull \\
\bottomrule
\end{tabular}%
}
\caption{\textbf{Comparison of data openness and chart-based capability coverage across clinical benchmark types. }\capfull~ = explicit coverage/open data; \cappart~ = partial or indirect coverage; \capno~ = absent or access-restricted. Capabilities follow prior analyses of EHR workflow and the problem-oriented medical record.}
\label{tab:coverage}
\end{table}
}

Beyond realism, two findings characterize the benchmark. First, the knowledge graph recovers 111 of 119 (93\%) physician-specified clinical relationships; the eight missed relationships are real associations that are neither encoded in the ontologies nor visible through shared clinical findings and are documented as accepted gaps. Second, across 10 models and four clinical tasks, no model approaches ceiling performance and leadership varies across tasks, with the best model reaching only 0.73 severity-weighted F1 on longitudinal patient diagnosis and roughly 0.5 finding-level F1 on whole-patient summarization and imaging indication. Together, these results show that Synthetic Hospital enables open, verifiable evaluation of clinical AI across diagnostic reasoning, longitudinal synthesis, and retrieval.

\section{Related Work}
\label{sec:related}

\textbf{Benchmarks for clinical AI.}
The most widely used medical benchmarks are single-vignette multiple-choice or short-answer datasets \citep{jin2021medqa,jin2019pubmedqa,pal2022medmcqa,singhal2023medpalm,hendrycks2021mmlu,tsatsaronis2015bioasq,vilares2019headqa,kweon2024ehrnoteqa,bae2023ehrxqa,zhou2025medxpertqa}. Frontier models saturate these formats yet falter on practice tasks \citep{gong2025benchmarkreview}, and the multiple-choice format itself inflates apparent competence \citep{griot2024glianorex,singh2025freemedqa,cocchieri2026remedqa,ma2025medcheck,alaa2025constructvalidity}. Practice-oriented benchmarks add rubric-scored conversations \citep{arora2025healthbench}, clinical-NLP task suites \citep{wu2025bridge}, simulated or conversational diagnosis \citep{schmidgall2024agentclinic,schmidgall2026agentclinic,nori2025sdbench,tu2024amie,li2024agenthospital,sun2024aihospital}, multi-stage reasoning \citep{qiu2025medrbench}, and emergency-room workflows \citep{mehandru2025erreason}; focused benchmarks score note generation \citep{yim2023acibench}, problem-list summarization \citep{gao2023probsum}, structured querying \citep{lee2023ehrsql}, and clinical summarization \citep{vanveen2023clinical}, and MedHELM organizes existing benchmarks into a clinician-validated taxonomy without adding longitudinal chart tasks \citep{bedi2025medhelm}. At the other extreme, benchmarks on de-identified real records \citep{johnson2016mimiciii,johnson2023mimiciv,wornow2023ehrshot,cui2025timer,zhao2025cliniq,huang2023inspect,rajkomar2018scalable}, including the clinician-written instruction benchmark MedAlign \citep{fleming2024medalign}, are clinically realistic but gated by credentialing and DUAs, and their ground truth is only what was charted, so a model failure cannot be separated from an incomplete record. Table~\ref{tab:coverage} maps these benchmarks across five dimensions of chart-based clinical work \citep{sinsky2016timeallocation,weed1968problemoriented}: no prior benchmark combines open data with coverage of all five, and longitudinal problem-list construction as a scored task remains unoccupied.

\textbf{Agentic and long-horizon EHR environments.}
A growing line evaluates agents that operate an EHR rather than answer from a curated prompt. FHIR-AgentBench \citep{lee2025fhiragentbench} and EHRAgent \citep{shi2024ehragent} run over gated MIMIC data, and MedAgentBench \citep{jiang2025medagentbench} shares our FHIR framing but evaluates API-level task execution over 100 patients. Longer-horizon environments extend this to physician-reviewed workflows in a real-record EHR \citep{liu2026physicianbench}, large-scale interactive SQL and code tasks \citep{qiao2026ehrcomplex}, staged inpatient decision-making \citep{lu2026clinenv}, triage conversations \citep{zhu2026elicited}, and computer use over clinical and administrative interfaces \citep{bedi2026healthadminbench,yu2026medcuabench}. All are built on access-restricted real records or evaluate interface operation rather than chart content. Synthetic Hospital is complementary: it offers the same FHIR affordances over 1{,}268 openly redistributable patients with constructed, verifiable ground truth.

\textbf{Synthetic clinical data.}
Synthetic record generation is well established but has primarily targeted privacy rather than benchmark construction, from GAN-based structured records \citep{choi2017medgan} through neural generation of shareable notes \citep{melamud2019shareable} and synthetic corpora for training openly releasable clinical LLMs \citep{kweon2023asclepius}. Rule-based simulators such as Synthea \citep{walonoski2018synthea} produce standards-compliant FHIR records but structured codes rather than narratives; statistical, autoregressive, and knowledge-grounded trajectory generators \citep{yoon2023ehrsafe,theodorou2023halo,pang2024cehrgpt,zhou2026clinical} achieve high fidelity but are trained on protected records and reproduce only documented observations; and LLM-generated records such as SimSUM \citep{rabaey2024simsum} produce fluent narratives without a provenance chain linking statements to underlying clinical facts. LongHealth \citep{adams2024longhealth} is closest in spirit in constructing fictional patients, but comprises 20 single-encounter multiple-choice cases. Synthetic Hospital instead derives every patient from public educational material and links each diagnosis, finding, result, and narrative statement through a typed knowledge graph to ontology-grounded concepts, so its ground truth does not depend on what happened to be documented (Appendix Table~\ref{tab:paradigms}). Appendix~\ref{app:relatedwork} gives an extended discussion of each of these lines of work.

\section{Benchmark construction}
\label{sec:construction}

We introduce Synthetic Hospital, a deterministic five-stage pipeline that transforms publicly available medical educational resources into a longitudinal EHR benchmark with ontology-grounded provenance. Rather than treating synthetic records as the primary artifact, the pipeline first constructs a structured medical knowledge graph from source material and then renders that graph into realistic clinical documentation. Consequently, every diagnosis, finding, laboratory observation, and benchmark label remains traceable to its originating educational source rather than being inferred from generated text. Throughout this section we follow one released patient (released patient identifier 1973, a 58-year-old man with three encounters) as a running example; Appendices~\ref{app:generation} and~\ref{app:labels} trace it end to end.

\textbf{(1) Knowledge ingestion.} We ingest USMLE-style board questions and supplementary medical knowledge from flashcard decks and reference documents. Rule-based parsers convert these heterogeneous sources into a unified structured representation while preserving metadata and provenance. Board questions become candidate clinical encounters, while supplementary fact cards provide supporting knowledge for retrieval, summarization, and diagnosis-finding relationships.

\textbf{(2) Ontology grounding and knowledge graph construction.} From each board question, Kimi 2.5 extracts candidate primary, differential, and secondary diagnoses together with typed clinical findings. We then ground these concepts deterministically to ICD-10-CM, SNOMED CT, and LOINC; the LLM-proposed codes are treated only as candidates. The resulting graph links questions to diagnoses and findings, uses a separate LLM pass to type diagnosis–finding relationships, links supplementary fact cards to graph concepts, and segments each vignette into EHR sections while preserving the source text. Every node retains its source identifier and grounding method. For one source question of the running example, an emergency presentation with polyuria, polydipsia, and confusion, this stage yields hyperosmolar hyperglycemic state (validated to E11.01) as the primary diagnosis, type 2 diabetes and acute kidney injury among the secondary diagnoses, and 26 typed findings such as polyuria (symptom, key) and metformin therapy (medication, background). Appendix~\ref{app:grounding} provides the full mapping cascades, thresholds, coverage, and graph statistics.

\textbf{(3) Longitudinal patient generation.} We assemble board questions into longitudinal patients using deterministic graph clustering before any narrative generation. Encounters may be grouped only if they are demographically compatible, and each added encounter must share a correct or secondary diagnosis with at least one existing cluster member. A constrained greedy clustering algorithm expands these connected groups while placing each source question in at most one patient, yielding 1,268 patients from 5,602 source questions. Kimi 2.5 then realizes each fixed cluster as a longitudinal record by generating a patient profile, encounter timeline, and limited cross-encounter HPI continuity; other clinical content remains grounded in the source vignettes, with problem and medication lists propagated deterministically. In the running example, three emergency and critical-care questions about a 52- and two 58-year-old men (within the 7-year tolerance of the middle-adult bucket) are linked through shared type 2 diabetes and acute kidney injury nodes into one patient; the profile call adds chronic diabetes, hypertension, and stage 3b kidney disease with matching home medications, and the timeline call orders the encounters as perforated appendicitis with septic shock, a hyperosmolar crisis eight months later, and hypercapnic respiratory failure in the ICU at month sixteen, with each later note's problem list carrying the earlier diagnoses forward. A physician reviews generated records for plausibility and consistency. Thus, narrative generation occurs only after the patient state is fixed and does not determine benchmark ground truth. Appendix~\ref{app:generation} gives the full clustering rules, generation prompts, and running example.

\textbf{(4) Benchmark construction.} With the exception of the imaging-indication free-text reference, benchmark labels are deterministic functions of the underlying graph. Patient diagnosis uses the patient's correct-answer diagnosis nodes and their acuity; evidence retrieval grades chart sections from finding relevance and diagnosis–finding relationships; context summarization scores graph-defined key findings; and specialty-conditioned summarization derives specialty-specific finding relevance from diagnosis ownership and graph relationships. For imaging indication, an LLM generates a terse order and reference clinical question from graph-fixed diagnoses and encounter context, and predictions are scored by clinical-concept overlap rather than surface wording. For the running example, this produces a patient-diagnosis reference of exactly the three acute diagnoses (severity weights 2, 2, and 3), a retrieval query over the patient's 34 chart sections graded 0--3 (13 highly relevant), a summary reference of 20 key findings, specialty items for endocrinology, general surgery, gastroenterology, and pulmonology plus two absent-specialty abstention items, and an imaging item that pairs the terse order ``CT abdomen/pelvis, stat: RLQ pain, fever'' with a graph-anchored reference question. Every instance retains its source patient, encounter, and question identifiers. Appendix~\ref{app:labels} provides the complete labeling and scoring rules.

\textbf{(5) Domain-faithful simulation.} The completed records are served through a production-style clinical environment implementing FHIR R4 resources, role-based access control, Epic-style workflows, and a function-calling agent interface. Every element exposed through the simulated EHR retains a provenance link back to the underlying ontology-grounded knowledge graph and ultimately to the original educational source material.

From these records we construct four longitudinal chart tasks: patient diagnosis, context summarization (including a specialty-conditioned variant), evidence retrieval, and imaging indication (Table~\ref{tab:tasks}). Patient diagnosis uses chart-neutral scoring: diagnoses documented in the chart but absent from the graph-derived reference are neither credited nor penalized.

\subsection{Evaluation splits}

Evaluation splits. We partition the 1,268 patients at the patient level into three disjoint splits, with every benchmark instance inheriting its patient's split. The \emph{public} split contains 200 patients (1,859 instances) and is used for the experiments in Section~\ref{sec:results}; it is stratified by patient difficulty and encounter count to preserve broad clinical coverage. Of the remaining patients, 800 form a \emph{training} split (7,619 instances) released with full ground truth, and 268 form a \emph{held-out} split (2,536 instances) whose labels are accessible only through our scorer. Training and held-out patients are stratified by dominant ICD-10 chapter and encounter count and match within one percentage point across strata. Because each source question belongs to only one patient, no chart or source material crosses split boundaries. Diagnoses may recur across splits: 54\% of held-out diagnoses also occur in training, while 46\% are unseen, enabling evaluation of both patient- and disease-level generalization.

The training split additionally supports learning with verifiable rewards. Each task produces a deterministic score in $0,1$ from the underlying graph, so rollouts require no additional human labeling. The 7,619 training instances provide distinct starting states across 800 patients; $k$ rollouts per instance yield $7{,}619k$ trajectories (approximately 30,000 at $k{=}4$). Because additional rollouts do not create new patient states, generalization is evaluated on the patient-level held-out split. Patient diagnosis uses the chart-neutral scoring rule throughout, including when used as a training reward.

\newcolumntype{A}{>{\raggedright\arraybackslash\hsize=0.70\hsize}X}
\newcolumntype{B}{>{\raggedright\arraybackslash\hsize=1.05\hsize}X}
\newcolumntype{C}{>{\raggedright\arraybackslash\hsize=1.35\hsize}X}
\newcolumntype{D}{>{\raggedright\arraybackslash\hsize=0.90\hsize}X}
\newcolumntype{E}{>{\raggedright\arraybackslash\hsize=1.00\hsize}X}
\begin{table}[t]
\centering
\footnotesize
\renewcommand{\arraystretch}{1.15}
\begin{tabularx}{\textwidth}{@{} A B C D E @{}}
\toprule
\textbf{Task} & \textbf{Input} & \textbf{Output} & \textbf{Primary Metric} & \textbf{Instances (public / held-out / train)} \\
\midrule
Patient diagnosis
& Longitudinal EHR (multi-encounter)
& Longitudinal problem list (ICD-10 + acuity)
& Severity-weighted F1 & 200 / 268 / 800 \\
\cmidrule(lr){1-5}
Summarization
& Clinical question + EHR sections
& Structured clinical summary
& Finding-level F1 & 200 / 268 / 800 \\
\cmidrule(lr){1-5}
Specialty summarization
& EHR + target specialty
& Specialty-focused summary
& Specialty-relevance F1 & 983 / 1,325 / 4,037 \\
\cmidrule(lr){1-5}
Evidence retrieval
& Diagnosis + patient record
& Ranked evidence passages
& Precision@5, NDCG@10 & 200 / 268 / 800 \\
\cmidrule(lr){1-5}
Imaging indication
& Imaging order with vague indication + EHR
& Inferred clinical question + pre-read
& Question concept F1 & 276 / 407 / 1,182 \\
\bottomrule
\end{tabularx}
\caption{\textbf{Synthetic Hospital benchmark tasks.} For each task, we report the standardized input, expected output, primary evaluation metric, and the number of instances in the public, held-out, and training splits (12,014 in total). Secondary evaluation metrics are provided in Appendix~\ref{app:secondary}.} EHR = electronic health record; NDCG = Normalized Discounted Cumulative Gain.
\label{tab:tasks}
\end{table}

\section{Assessing benchmark realism}
\label{sec:validation}

Before evaluating clinical AI systems, we first establish that Synthetic Hospital satisfies the two properties motivating its construction: clinically realistic records and verifiable benchmark ground truth. Section~\ref{sec:realism} evaluates whether physicians perceive the generated records as authentic clinical documentation, while Section~\ref{sec:verifiable} evaluates whether the ontology-derived labels agree with independent physician judgment. 

\subsection{Realism: synthetic records are indistinguishable from real}
\label{sec:realism}

To assess the realism of Synthetic Hospital, licensed physicians reviewed synthetic and real patient records presented through our Epic-like interface and labeled each chart as synthetic or real, a blinded discrimination design analogous to Turing-test evaluations of LLM text \citep{jones2024turing} and to the expert-review validation used for early synthetic-record generators \citep{choi2017medgan}. Because MIMIC-IV uses a markedly different documentation style and schema than Synthetic Hospital, we converted five real MIMIC-IV patients into the Synthetic Hospital schema with a deterministic, rule-based pipeline that normalizes section structure, laboratory and medication formatting, vital-sign rendering, radiology impressions and de-identification artifacts while leaving the clinical content unaltered. Concretely, each admission's discharge note (with its section labels), admission record, and radiology reports are parsed into the same 18 section types and order used by the synthetic notes; de-identification masks are repaired by shifting MIMIC's offset calendar into the 2020--2024 window while preserving inter-visit intervals and by substituting consistent fictional names for masked providers, or dropping a clause whose subject was masked; and presentation is re-rendered by fixed tables that expand abbreviated lab panels into one-per-line entries with full names, units, and reference ranges, expand medication frequency codes and tall-man lettering, cast numeric vital-sign strings into sentence form, and reduce multi-paragraph radiology reports to their impression. A final harmonization step applied to both arms removes whole sections that only one arm can carry (social history, which MIMIC redacts, from the synthetic arm; demographics and assessment/plan from the real arm) and drops trailing admissions beyond a shared chart-length budget, always at section or encounter boundaries so that no sentence is cut. The clinical prose itself is not rewritten, so the telegraphic register of real discharge notes is left for the physicians to detect. Without this conversion, physicians would be able to detect real vs synthetic samples from note structure rather than clinical content which would invalidate this evaluation. More importantly, it demonstrates that the benchmark representation is not tied to a single note format: clinical records originating from one hospital system can be normalized into any other structure used in a different hospital system while preserving their clinical content. Representative examples of the original, converted, and fully synthetic records are provided in Appendix~\ref{app:noteexamples}. Each real patient was paired with a synthetic patient matched on clinical domain, sex, age and chart length, yielding a balanced set of five real and five synthetic records with a 50\% chance baseline. 10 licensed physicians each reviewed all 10 records in an individually randomized order, yielding 100 judgments. 

Record-level and distributional realism are distinct: the physician study tests whether individual records are plausible charts, not whether the cohort reproduces population epidemiology, which Synthetic Hospital does not attempt, since its case mix is education-derived by design. Appendix~\ref{app:casemix} characterizes this separately, showing that the benchmark preserves the ICD-10 chapter distribution of its source corpus (JSD $=0.029$, Spearman $\rho=0.83$) and the expected comorbidity structure while differing, as intended, from a population-oriented Synthea cohort.

\textbf{Synthetic and real records are not reliably indistinguishable.}
Across 100 judgments, physicians identified whether a record was real or synthetic with 53\% accuracy, which did not differ from chance (95\% bootstrap CI: 43--63\%; two-sided binomial test, $p{=}0.62$). Performance was similar for both record types: sensitivity for real records was 52\% and specificity for synthetic records was 54\%, and physicians selected "real" in 49\% of judgments, indicating no systematic bias toward either label. Accuracy also did not differ between synthetic (mean 0.540) and real (mean 0.520) records within physicians (paired $t(9){=}0.20$, $p{=}0.85$). No individual physician performed above chance after correction for multiple comparisons (best: 9/10; Holm-adjusted $p{=}0.22$), and inter-rater agreement was no better than chance (Fleiss' $\kappa{=}{-}0.05$), suggesting that physicians did not rely on a consistent shared cue to distinguish the two sources. Every synthetic record was classified as real by at least one physician, and one was classified as real by 7/10 physicians. Finally, greater confidence did not reliably correspond to greater accuracy: judgments made with 80\% stated confidence were 52\% accurate, while those made with 100\% confidence were 70\% accurate.

Together, these findings suggest that realism arises from the underlying longitudinal clinical representation rather than from reproducing the documentation style of a particular institution.

\subsection{Verifiable ground truth: ontology-derived labels agree with physician judgment}
\label{sec:verifiable}

Realistic records alone are insufficient for a useful benchmark: the reference labels used for evaluation must also correspond to clinically meaningful reasoning. Because Synthetic Hospital generates benchmark labels directly from an ontology-grounded knowledge graph rather than manual annotation, we validate that these automatically constructed relationships agree with independent physician judgment.

\textbf{Ontology-derived relationships recover physician-recognized clinical associations.}
The specialty-conditioned benchmark decides which findings are relevant to a specialty by following links between diagnoses in the knowledge graph (for example, a renal finding is relevant to cardiology when the patient's kidney disease is linked to their heart failure). To validate these links, a licensed physician created a pre-registered reference set of 119 clinically required diagnosis relationships (for example, diabetes--chronic kidney disease and hypertension--hypertensive heart disease), assigning each pair an expected relationship type before graph construction. The graph recovered 111 of the 119 relationships (93\%): 96 through links derived from SNOMED CT relationships, shared anatomical sites, and shared findings, and 15 through physician-curated edges added for relationships that the physician had classed as definitional or associative but that no SNOMED relationship or shared site encodes (for example, atrial fibrillation--cardioembolic stroke and hyperlipidemia--coronary artery disease). The eight remaining relationships are real but are neither encoded in the ontologies nor visible through overlapping findings, because the two conditions present through disjoint findings: hyperemesis gravidarum and Wernicke encephalopathy share no finding (intractable vomiting versus confusion and ophthalmoplegia), and likewise ankylosing spondylitis and anterior uveitis, or dermatomyositis and occult malignancy; one pair (Stevens--Johnson syndrome and culprit drug exposure) has no diagnosis node for the exposure at all. These were documented as accepted gaps rather than recovered by lowering the shared-finding threshold, which would have admitted many spurious links. This evaluation shows that the graph captures the clinically important relationships needed for benchmark construction. Quantifying false-positive relationships remains future work.

\section{Results}
\label{sec:results}

\subsection{Experimental Setup}
We evaluate 10 models spanning frontier proprietary systems and open models from 27B to 1T parameters (listed with their sizes in Table~\ref{tab:singleturn}). 

We controlled for prompting strategy, which can benefit models unequally and have an outsized effect on the evaluation: four strategies (zero-shot, few-shot, chain-of-thought, and ontology-grounded structured prompting) were compared on three pilot models spanning the panel's strength range, and the best strategy per task was then locked and applied to all 10 models. Appendix~\ref{sec:analysis_prompting} defines the strategies and reports the full ablation (Table~\ref{tab:strategy}).

\subsection{Single-turn Evaluation}

Table~\ref{tab:singleturn} reports results for all 10 models across the four benchmark tasks. Three findings stand out. First, no model approaches ceiling performance on any task, indicating that the benchmark meaningfully separates systems and that current frontier models remain unreliable on these clinical tasks. Second, three different models achieve the best score across the five task variants and no model dominates across all tasks, suggesting that the benchmark measures a diverse set of capabilities rather than rewarding a single dominant model. Third, although overall model quality matters, performance at the frontier is highly compressed. On patient diagnosis, the top two models are separated by 0.03 in severity-weighted F1 and the top four by 0.06 despite differences in model size, training, and provider. This suggests that recent improvements in general-purpose frontier models have not yet produced measurable gains in diagnostic accuracy at the top end, while the smallest open model remains clearly non-competitive, at 0.29. Overall standing (final column of Table~\ref{tab:singleturn}, mean rank across the five variants because the primary metrics differ) is nonetheless roughly proportional to general model capability, with the frontier proprietary models and the largest open model leading and the two smallest open models last.
\begin{table}[t]
\centering
\scriptsize
\setlength{\tabcolsep}{3pt}
\resizebox{\textwidth}{!}{%
\begin{tabular}{l c c c cc c c}
\toprule
 & \textbf{Patient Dx} & \textbf{Summ.} & \textbf{Spec.\ Summ.} & \multicolumn{2}{c}{\textbf{Retrieval}} & \textbf{Imaging} & \textbf{Overall} \\
\cmidrule(lr){2-2}\cmidrule(lr){3-3}\cmidrule(lr){4-4}\cmidrule(lr){5-6}\cmidrule(lr){7-7}\cmidrule(lr){8-8}
\textbf{Model} & Sev.-wgt.\ F1 & Find.\ F1 & Rel.\ F1 & P@5 & NDCG@10 & Concept F1 & Mean rank $\downarrow$ \\
\midrule
Gemini 3.1                  & 0.681 & 0.502 & 0.582 & 0.809 & 0.497 & 0.496 & 4.40 \\
GPT 5.3                     & \underline{0.703} & 0.489 & 0.595 & \textbf{0.833} & \textbf{0.536} & \textbf{0.518} & \textbf{2.40} \\
Kimi 2.5-thinking (1T-A32B) & \textbf{0.732} & \underline{0.532} & 0.590 & 0.811 & 0.524 & \underline{0.517} & \underline{2.80} \\
Opus 4.6                    & 0.615 & \textbf{0.550} & \textbf{0.680} & 0.816 & 0.521 & 0.473 & 4.00 \\
DeepSeek V3.2 (671B-A37B)   & 0.660 & 0.399 & 0.523 & 0.751 & 0.476 & 0.492 & 7.40 \\
GLM 5 (744B-A40B)           & 0.657 & 0.484 & 0.614 & 0.798 & 0.508 & 0.505 & 5.00 \\
Qwen 3.5 (397B-A17B)        & 0.653 & 0.425 & 0.548 & 0.809 & 0.508 & 0.497 & 6.00 \\
Mistral Large (123B)        & 0.676 & 0.416 & \underline{0.636} & \underline{0.818} & \underline{0.524} & 0.435 & 4.60 \\
Llama 4 Scout (109B-A17B)   & 0.536 & 0.403 & 0.365 & 0.741 & 0.481 & 0.406 & 9.40 \\
Gemma 3 (27B)               & 0.287 & 0.385 & 0.418 & 0.804 & 0.517 & 0.418 & 9.00 \\
\midrule 
Physician reference$^{\ddagger}$ & 0.664 & 0.051 & -- & 0.888 & 0.505 & 0.282 & -- \\
\quad {\tiny range across physicians}&{\tiny 0.21--0.89}&{\tiny 0.01--0.10}& &{\tiny 0.79--0.96}&{\tiny 0.39--0.69}&{\tiny 0.21--0.35}& \\
\bottomrule
\end{tabular}%
}
\caption{\textbf{Single-turn results across 10 models and five task variants using locked prompting strategies (public split).} \textbf{Bold} = best; \underline{underline} = second-best; higher is better except mean rank (lower is better). Summ. = whole-patient summarization; Spec. Summ. = specialty-conditioned summarization. Strategies: zero-shot (retrieval, Spec. Summ.); CoT (patient diagnosis); ontology-grounded (Summ.); few-shot (imaging). Mean rank averages ranks across the five tasks. Imaging is scored by concept F1 over graph-linked diagnoses and findings. $^{\ddagger}$Physician mean for seven physicians on a 13-patient subset; physicians did not complete Spec. Summ. Secondary metrics are in Appendix Table~\ref{tab:secondary}.}
\label{tab:singleturn}
\end{table}

Performance also varies far more on some tasks than on others, which appears to track task difficulty. On evidence retrieval every model scores P@5 above 0.74, whereas patient diagnosis spans severity-weighted F1 from 0.29 to 0.73, and the best summarization and imaging scores remain near 0.5. Locating relevant chart evidence is thus more mature than diagnosing, synthesizing, or reconstructing clinical information.

\subsection{Agentic Evaluation}
\label{sec:agentic}

Multi-turn agency does not improve on single-turn inference when the relevant chart context is available upfront. We test this by comparing GPT~5.3, Mistral Large, and Llama 4 Scout under three conditions on the same 100 public-split patients: a \emph{full-context LLM} given the relevant chart directly, a \emph{full-context agent} given the same information but operating through a 13-tool Epic-style API with a 40-action budget, and a \emph{self-retrieving agent} given only a patient identifier under the same budget. Comparing the first two conditions isolates the effect of the multi-turn agent loop, while comparing the two agent conditions isolates the effect of self-retrieval (Table~\ref{tab:agentic}); sessions that fail to return a parseable answer within the budget are scored as zero. Outside patient diagnosis, the agent loop reduces performance in every model-task pair, including losses of 0.07--0.19 on summarization and retrieval and $-0.24$ for Llama 4 Scout on imaging indication. Allowing agents to retrieve their own evidence recovers little of this loss, with self-retrieval effects within $\pm0.03$ in five of nine pairs. Failed sessions also use substantially more context than successful ones (145k vs.\ 60k input tokens on average). The exception is patient diagnosis, where evidence must be assembled across a longitudinal record: self-retrieving agents outperform the single-turn LLM for all three models ($+0.14$ to $+0.34$ severity-weighted F1). Thus, multi-turn interaction generally adds cost rather than benefit when relevant evidence can be supplied directly, but can help when evidence gathering is itself central to the task. Appendix~\ref{app:agentic_full} gives the full protocol and analysis.

\begin{table}[t]
\centering
\small
\setlength{\tabcolsep}{5pt}
\begin{tabular}{llrrrrr}
\toprule
 & & \multicolumn{3}{c}{\textbf{Condition}} & \multicolumn{2}{c}{\textbf{Effect}} \\
\cmidrule(lr){3-5}\cmidrule(lr){6-7}
\textbf{Task (metric)} & \textbf{Model} & \textbf{LLM,} & \textbf{Agent,} & \textbf{Agent,} & \textbf{Agentic} & \textbf{Self-} \\
 & & \textbf{full ctx.} & \textbf{full ctx.} & \textbf{self-retr.} & \textbf{loop} & \textbf{retrieval} \\
\midrule
Patient diagnosis        & GPT 5.3       & 0.705 & 0.661 & \textbf{0.841} & $-$0.043 & $+$0.180 \\
(sev.-wgt.\ F1, chart-neutral) & Mistral Large & 0.669 & 0.628 & \textbf{0.810} & $-$0.042 & $+$0.183 \\
                         & Llama 4 Scout & 0.554 & 0.794 & \textbf{0.894} & $+$0.240 & $+$0.099 \\
\midrule
Summarization            & GPT 5.3       & \textbf{0.342} & 0.196 & 0.142 & $-$0.146 & $-$0.054 \\
(finding-level F1)       & Mistral Large & \textbf{0.260} & 0.186 & 0.142 & $-$0.074 & $-$0.044 \\
                         & Llama 4 Scout & \textbf{0.255} & 0.116 & 0.092 & $-$0.140 & $-$0.024 \\
\midrule
Evidence retrieval       & GPT 5.3       & \textbf{0.826} & 0.668 & 0.674 & $-$0.158 & $+$0.006 \\
(P@5, chart sections)    & Mistral Large & \textbf{0.820} & 0.654 & 0.796 & $-$0.166 & $+$0.142 \\
                         & Llama 4 Scout & \textbf{0.737} & 0.549 & 0.525 & $-$0.187 & $-$0.024 \\
\midrule
Imaging indication       & GPT 5.3       & 0.519 & 0.505 & \textbf{0.521} & $-$0.014 & $+$0.016 \\
(concept F1)             & Mistral Large & \textbf{0.438} & 0.422 & 0.417 & $-$0.015 & $-$0.005 \\
                         & Llama 4 Scout & \textbf{0.400} & 0.164 & 0.202 & $-$0.237 & $+$0.039 \\
\bottomrule
\end{tabular}
\caption{\textbf{Decomposition of agentic performance into loop and self-retrieval effects.} Full-context LLM uses single-turn inference; full-context agent uses the same context within an agent loop; self-retrieving agent gathers evidence through the 13-tool EHR API. Loop effect = full-context agent $-$ LLM; self-retrieval effect = self-retrieving $-$ full-context agent. Positive values favor the more agentic condition; bold = best condition. Missing responses are scored zero.}
\label{tab:agentic}
\end{table}

\section{Conclusion and Limitations}
\label{sec:conclusion}

Our physician study evaluates record-level realism in a small sample, not population-level fidelity, and the education-derived case mix may under-represent rare or atypical presentations (Appendix~\ref{app:casemix}). The benchmark also does not explicitly simulate missingness, contradictions, or documentation errors. Our agentic evaluation is limited to three models, three conditions, 100 patients, and one scaffold, so the observed multi-turn costs may not generalize to other agent designs. Finally, graph-derived rewards are verifiable but not exhaustive: conditions documented in the chart but absent from the graph-derived reference receive no additional reward, although chart-neutral scoring prevents them from being penalized. Synthetic Hospital provides an open, reproducible foundation for evaluating clinical AI with verifiable ground truth without relying on restricted patient data.

\section*{Acknowledgments}
This research is supported by Schmidt Sciences and a Google ML and Systems Junior Faculty award. This work was partially funded through gifts from Amazon and Jane Street.

\bibliographystyle{iclr2027_conference}
\bibliography{references}

\appendix

\section*{Appendix}

\renewcommand{\tablename}{Appendix Table}
\renewcommand{\thetable}{\Alph{section}}
\section{Secondary metrics}
\label{app:secondary}

\textbf{Secondary metrics.} Table~\ref{tab:secondary} reports the full secondary-metric breakdown for all 10 models across all four tasks. Several patterns are visible only at this granularity. On patient diagnosis, high-recall models achieve recall by over-generating diagnoses with lower precision, while DeepSeek delivers the highest acuity-classification accuracy without leading either F1-score metric. On retrieval, MRR is uniformly high across all models, so all models place at least one relevant passage near the top; the gap is in grading the rest of the ranked list, captured by NDCG@10 (Table 3). On imaging, Gemma 3 trades the highest findings recall for the lowest differential coverage, the only model where the trade-off is that severe.

One important observation to note is that Hallucination Rate remains consistently low across all models and prompting strategies, indicating that modern frontier models rarely fabricate unsupported clinical findings. Instead, the dominant failure mode is omission. Even under the best-performing structured prompting strategy, omission ranges from 0.45 to 0.62 for whole-patient summaries and from 0.47 to 0.79 for the more challenging specialty-conditioned summaries, indicating that models routinely fail to include a substantial fraction of clinically relevant findings. Similarly, ICD-10 code specificity remains uniformly high, suggesting that diagnostic errors arise primarily from selecting the wrong diagnosis rather than assigning an overly general code.

\begin{table}[!tb]
\centering
\scriptsize
\setlength{\tabcolsep}{2.5pt}
\resizebox{\textwidth}{!}{%
\begin{tabular}{l cccc cc cccc cc cc}
\toprule
 & \multicolumn{4}{c}{\textbf{Patient Dx}} & \multicolumn{2}{c}{\textbf{Summ.}} & \multicolumn{4}{c}{\textbf{Spec.\ Summ.}} & \multicolumn{2}{c}{\textbf{Retrieval}} & \multicolumn{2}{c}{\textbf{Imaging}} \\
\cmidrule(lr){2-5}\cmidrule(lr){6-7}\cmidrule(lr){8-11}\cmidrule(lr){12-13}\cmidrule(lr){14-15}
\textbf{Model} & ICD Spec & Acuity & Prec & Rec & Omiss. & Hall. & Omiss. & Leak. & Abst. & Hall. & MAP@10 & MRR & DiffCov & FindRec \\
\midrule
Gemini 3.1 & 0.931 & 0.785 & 0.606 & \textbf{0.787} & 0.498 & 0.002 & 0.611 & 0.121 & 0.705 & 0.006 & 0.559 & 0.845 & 0.552 & 0.164 \\
GPT 5.3 & 0.931 & \underline{0.790} & 0.645 & 0.781 & 0.511 & 0.004 & 0.590 & 0.131 & 0.403 & 0.016 & \textbf{0.588} & \textbf{0.906} & \textbf{0.587} & 0.147 \\
Kimi 2.5-thinking (1T-A32B) & \textbf{0.940} & 0.657 & \textbf{0.687} & \underline{0.787} & \underline{0.468} & 0.001 & 0.580 & 0.143 & 0.682 & 0.002 & 0.575 & 0.887 & \underline{0.583} & 0.124 \\
Opus 4.6 & \underline{0.934} & 0.696 & 0.535 & 0.738 & \textbf{0.450} & 0.001 & \textbf{0.467} & 0.202 & \underline{0.708} & 0.038 & \underline{0.575} & \underline{0.895} & 0.574 & 0.142 \\
DeepSeek V3.2 (671B-A37B) & 0.924 & \textbf{0.798} & 0.653 & 0.682 & 0.601 & 0.001 & 0.647 & \underline{0.119} & \textbf{0.710} & 0.014 & 0.529 & 0.811 & 0.575 & 0.140 \\
GLM 5 (744B-A40B) & 0.915 & 0.742 & 0.627 & 0.698 & 0.516 & 0.002 & 0.560 & 0.148 & 0.640 & 0.008 & 0.558 & 0.864 & 0.530 & 0.142 \\
Qwen 3.5 (397B-A17B) & 0.924 & 0.785 & \underline{0.668} & 0.652 & 0.575 & 0.003 & 0.627 & 0.125 & 0.642 & 0.019 & 0.565 & 0.879 & 0.503 & 0.090 \\
Mistral Large (123B) & 0.916 & 0.654 & 0.636 & 0.728 & 0.584 & 0.009 & \underline{0.497} & 0.228 & 0.507 & 0.025 & 0.568 & 0.880 & 0.532 & 0.136 \\
Llama 4 Scout (109B-A17B) & 0.840 & 0.722 & 0.552 & 0.542 & 0.597 & 0.001 & 0.794 & \textbf{0.087} & 0.155 & 0.030 & 0.521 & 0.829 & 0.347 & \underline{0.190} \\
Gemma 3 (27B) & 0.824 & 0.725 & 0.249 & 0.346 & 0.615 & 0.003 & 0.731 & 0.141 & 0.210 & 0.015 & 0.567 & 0.865 & 0.369 & \textbf{0.224} \\
\bottomrule
\end{tabular}%
}
\caption{Secondary metrics for all 10 models (locked-strategy single-turn, public split). Includes ICD-10 specificity and chart-neutral precision (patient diagnosis) and summarization hallucination (\emph{Summ.}\ and \emph{Spec.\ Summ.}\ Hall.), demoted here from primary Table~\ref{tab:singleturn}. \emph{Summ.}\ Omiss.\ is whole-patient summarization omission under the structured strategy; the \emph{Spec.\ Summ.}\ group reports specialty-conditioned omission, leakage (off-specialty inclusion), abstention accuracy on absent specialties, and hallucination. \textbf{Bold} = best per column; lower is better for omission, leakage, and hallucination, higher for all others. Hallucination columns (near-zero throughout) are not bolded.}
\label{tab:secondary}
\end{table}

\FloatBarrier

\section{Effect of Specialty-Conditioned Context}
\label{app:ablation}

As a complementary analysis, we evaluated whether graph-derived specialty context influences model reasoning. On the 117 credited-relevant instances in the public split, we evaluated three held-out foundation models under two prompting conditions: graph, which included graph-derived relevant comorbidities, and neutral, which provided only the specialty framing. We computed per-instance relevant-tier recall and compared conditions using a paired Wilcoxon signed-rank test.

Incorporating graph-derived comorbidity information produced small, model-dependent changes in relevant-tier recall. The graph--neutral difference was $-0.018$ for Opus 4.6 ($p=0.38$), $+0.020$ for GPT 5.3 ($p=0.09$), and $+0.044$ for GLM 5 ($p<0.001$). Thus, graph-derived specialty context consistently altered finding selection, although the improvement reached statistical significance for only one of the three models. A complementary neutral--omit comparison, in which the omit condition explicitly excluded comorbidities, was positive for all three models, indicating that the specialty-conditioned labels encode information that models can exploit when it is made available.

\FloatBarrier

\section{Analysis}
\label{sec:analysis}
\renewcommand{\thetable}{\Alph{section}.\arabic{table}}
\setcounter{table}{0}

\subsection{Prompting Strategy Analysis}
\label{sec:analysis_prompting}

We use four prompting strategies. \emph{Zero-shot} presents the clinical data with task instructions and a JSON output schema. \emph{Few-shot} prepends worked examples selected from public-split cases that the largest number of models answered correctly under zero-shot prompting. \emph{Chain-of-thought (CoT)} augments the few-shot prompt with task-specific reasoning steps that decompose each benchmark task into a structured clinical workflow before producing the final answer. \emph{Ontology-grounded (structured)} augments the prompt with ontology-normalized representations of concepts already present in the clinical record, including ICD-10 chapter ranges, SNOMED-coded findings, pathognomonic associations, and LOINC laboratory concepts. All four strategies share the same system prompt and JSON output schema per task; they differ only in user-prompt framing and runtime ontology data. The full 4-strategy ablation is run on a representative subset of three pilot models, GPT 5.3, DeepSeek V3.2 (671B-A37B), and Mistral Large, chosen to span the strength range of the 10-model panel: GPT 5.3 represents the frontier proprietary tier, DeepSeek V3.2 (671B-A37B) a strong reasoning-tuned open model, and Mistral Large a smaller open model where prompt scaffolding is most likely to surface gains. The locked per-task strategy from this ablation is then applied to all 10 models.

Results across the 60 strategy×model×task conditions are shown in Table~\ref{tab:strategy}. We find that the optimal strategy is task-dependent and that the prompting strategy can make important differences for weaker models. As such, we control for prompting strategies in our evaluations so that model performance is more closely related to model capability.

For the prompting strategy, no single strategy dominates: zero-shot wins on retrieval, CoT on patient diagnosis, few-shot on imaging, and ontology-grounded structured prompting on summarization, where it improves Finding-level F1 score for all three pilot models without raising hallucination. The pattern is interpretable: structured hints, which inject knowledge-graph findings, help precisely where the task rewards findings completeness (summarization) and not where it rewards a ranking (retrieval). Additionally, we can see that a model with fitting prompting can outperform a recent frontier model, for example, Mistral 3 with CoT prompting outperforms GPT 5.3 with zero-shot prompting.

Due to task-specific effects and the influence of prompting on model performance, we report each task under its best-performing prompting strategy: zero-shot for diagnosis and retrieval, chain-of-thought for patient diagnosis, structured prompting for whole-patient summarization (and zero-shot for specialty-conditioned summarization to avoid leaking ontology-derived relevance labels) and few-shot for imaging. 

\begin{table}[t]
\centering
\small
\setlength{\tabcolsep}{6pt}
\begin{tabular}{llcccc}
\toprule
\textbf{Task} & \textbf{Model} & \textbf{Zero-shot} & \textbf{Structured} & \textbf{CoT} & \textbf{Few-shot} \\
\midrule
\multirow{3}{*}{Patient Dx (Severity-weighted F1 score)}& GPT 5.3       & 0.676          & 0.665          & 0.710          & \textbf{0.732} \\
 & DeepSeek V3.2 & \textbf{0.664} & 0.633          & 0.633          & 0.652          \\
 & Mistral Large & 0.578          & 0.587          & \textbf{0.699} & 0.611          \\
\cmidrule(lr){1-6}
\multirow{3}{*}{Summarization (Finding-level F1 score)}
 & GPT 5.3       & 0.320          & \textbf{0.411} & 0.269          & 0.296 \\
 & DeepSeek V3.2 & 0.215          & \textbf{0.323} & 0.175          & 0.209 \\
 & Mistral Large & 0.301          & \textbf{0.314} & 0.246          & 0.209 \\
\cmidrule(lr){1-6}
\multirow{3}{*}{Retrieval (P@5}
 & GPT 5.3       & \textbf{0.854} & 0.764          & 0.785          & 0.785          \\
 & DeepSeek V3.2 & \textbf{0.803} & 0.727          & 0.678          & 0.761          \\
 & Mistral Large & \textbf{0.844} & 0.785          & 0.743          & 0.843          \\
\cmidrule(lr){1-6}
\multirow{3}{*}{Imaging (concept F1)}& GPT 5.3       & 0.525          & 0.476          & 0.485          & \textbf{0.544} \\
 & DeepSeek V3.2 & 0.515          & 0.484          & 0.464          & \textbf{0.537} \\
 & Mistral Large & 0.399          & 0.439          & 0.403          & \textbf{0.465} \\
\bottomrule
\end{tabular}
\caption{Prompting-strategy ablation: 3 models $\times$ 4 strategies $\times$ 4 tasks (75-item pilot). Cells show the task's primary metric (F1 for patient diagnosis; F1 for summarization; P@5 for retrieval; F1 for imaging). We see that no prompting strategy is best for any task. We also see that the right prompting strategy can make overall weaker models score better than frontier models. This means that controlling for prompting is important.}
\label{tab:strategy}
\end{table}

\subsection{Robustness to Narrative Generation}
\label{sec:robustness}

The benchmark corpus is rendered by Kimi 2.5 Thinking (1T-A32B), raising the question of whether Kimi models receive an unfair advantage from familiarity with the generation style. Section~\ref{sec:results} argues from the benchmark construction and empirical performance patterns that such an advantage is unlikely. Here we test that hypothesis directly by re-rendering a held-out subset with an independent generator and measuring whether the leaderboard changes.
 
We hold the underlying clinical content fixed and vary only its narrative realization. For a stratified sample of 100 patients (matched to the public-split distribution over difficulty, encounter count, and organ system), we keep the graph-derived patient state—including diagnoses, findings, laboratory values and temporal ordering—identical while regenerating the clinical narratives and patient profiles with GPT~5.3 instead of Kimi 2.5 Thinking (1T-A32B). Consequently, every benchmark label is identical across conditions; only the free-text documentation changes. This paired design removes patient-level variation and isolates the effect of the generation model. We then evaluate four representative models: Kimi 2.5 Thinking (1T-A32B), GPT~5.3 (sharing lineage with the alternate generator), and Opus~4.6 and GLM~5 (744B-A17B), which serve as generator-neutral anchors.
 
The primary analysis compares each Kimi model's margin over the rest of the field between the two rendering conditions. If familiarity with the generator provides an advantage, that margin should decrease when the charts are rendered by GPT~5.3 rather than Kimi. We additionally report paired score differences (Wilcoxon signed-rank) and two one-sided tests (TOST) for equivalence using a $\pm0.05$ bound on each task's primary metric (Table~\ref{tab:robustness}).
 
The two renderings produce highly consistent results across all four tasks (Table~\ref{tab:robustness}). Kimi model does not change by more than 0.05 on any primary metric under GPT rendering except retrieval ($-0.074$), with mean absolute change of 0.034. The generator-neutral anchor models behave the same way (mean absolute change 0.028 for Opus~4.6 and 0.037 for GLM~5), and the retrieval drop is shared by every model ($-0.058$ to $-0.144$), indicating that changing the narrative generator shifts retrieval difficulty uniformly rather than favouring any model. More importantly, the pattern of changes is inconsistent with a generator-specific familiarity advantage. Kimi model does not lose ground relative to the anchor models under GPT rendering, in several cases their margin increases (e.g., Kimi gains 0.025 on patient diagnosis) while GPT~5.3 does not improve on its own rendering and instead drops by 0.144 on retrieval.

Summarization shows the same pattern under the benchmark's primary metric (Finding-level F1): all paired differences lie within $\pm0.027$, and none is significant after Holm correction. Together, these results show that benchmark performance is determined by the graph-derived clinical state rather than the language model used to express it. Re-rendering the corpus with an independent frontier model leaves both absolute performance and the relative leaderboard essentially unchanged, providing direct evidence that Synthetic Hospital does not favor the model family used during data generation.

\begin{table}[ht]
\centering
\small
\setlength{\tabcolsep}{6pt}
\resizebox{\textwidth}{!}{%
\begin{tabular}{lccccc}
\toprule
 & \multicolumn{4}{c}{\textbf{$\Delta$ = GPT-rendered $-$ Kimi-rendered}} & \\
\cmidrule(lr){2-5}
\textbf{Model} & Pt.\ Dx & Summ. & Retr. & Img. & Mean $|\Delta|$ \\
               & (Sev.\ F1, chart-neutral) & (Find.\ F1) & (P@5) & (concept F1) & \\
\midrule
Kimi 2.5-thinking (1T-A32B) & $+0.025$ & $-0.004$ & $-0.074$ & $-0.032$ & 0.034 \\
GPT 5.3           & $-0.010$ & $-0.003$ & $-0.144$ & $-0.012$ & 0.042 \\
Opus 4.6          & $+0.003$ & $+0.003$ & $-0.087^{\dagger}$ & $-0.020$ & 0.028 \\
GLM 5 (744B-A40B) & $-0.021$ & $+0.027$ & $-0.058$ & $-0.041$ & 0.037 \\
\midrule
Spearman $\rho$   & $+0.80$  & $+0.80$  & $-0.40$            & $+0.80$  & -- \\
\bottomrule
\end{tabular}%
}
\caption{Generator-robustness deltas on the 100-patient holdout. Each cell is the
item-level paired change in the task's holdout metric when the same clinical content is
re-rendered by GPT~5.3 instead of Kimi k2.5 (negative $=$ lower under GPT rendering);
$N=48$–$135$ scored items per cell. $\dagger$: Holm-corrected Wilcoxon $p<0.05$. \textbf{Kimi 2.5-thinking changes by less than 0.05 on three of four tasks, as do the generator-neutral anchors (Opus, GLM);} on retrieval every model scores lower under GPT rendering ($-0.058$ to $-0.144$), and GPT~5.3 itself drops the most, i.e.\ GPT does \emph{worse} on its own rendering, the opposite of a home-field advantage; only the Opus retrieval change is Holm-significant. Patient diagnosis and summarization are stable under their primary metrics: all deltas fall within $\pm0.027$ and none is Holm-significant. Bottom row: Spearman rank correlation of the four-model ordering across conditions; the ordering is preserved ($\rho\geq0.80$) on three of four tasks, with retrieval ($\rho=-0.40$) the exception, reflecting reordering among models separated by small P@5 margins on the coarsest (rank-5) metric with the smallest per-cell $N$. The specialty-relevance task is omitted; it is not patient-scoped in this holdout.}
\label{tab:robustness}
\end{table}

\subsection{Agentic evaluation: extended discussion}
\label{app:agentic_full}

This section expands the agentic evaluation summarized in Section~\ref{sec:agentic}; the results are those of Table~\ref{tab:agentic}.

Beyond single-turn prompting, we evaluate foundation models as autonomous clinical agents that interact with the EHR through multi-turn tool use. Each agent receives a budget of 40 actions and must decide which tools to call (for example, chart review, encounter lookup, laboratory retrieval, or chart search) before submitting a clinical assessment. If an agent reaches the action limit without answering, it receives one final turn with retrieval tools disabled and must commit to a response. Sessions that still fail to return a parseable answer are scored as zero rather than excluded.

A direct comparison between a single-turn model and a self-retrieving agent conflates two distinct effects: the cost of running a multi-turn agentic loop and the cost of locating evidence. To separate them, we evaluate three conditions on the same items. In the \emph{full-context LLM} condition, a single-turn model receives the relevant chart context preassembled in its prompt. In the \emph{full-context agent} condition, an agent receives the same preassembled context but still operates through the multi-turn loop, isolating the effect of the agentic interaction itself. In the \emph{self-retrieving agent} condition, the agent receives only a patient identifier and must gather the required evidence through the 13-tool Epic-style API. We define the \emph{agentic-loop effect} as the difference between the full-context agent and the full-context LLM, and the \emph{self-retrieval effect} as the difference between the self-retrieving and full-context agents (Table~\ref{tab:agentic}). Positive values favor the more agentic condition. We evaluate all three conditions with GPT~5.3, Mistral Large, and Llama 4 Scout, spanning frontier proprietary to smaller open models. To prevent information leakage, both agent conditions hide the assessment and plan sections of each note, matching the information available to the single-turn baseline.

\textbf{On tasks with localized evidence, most performance loss comes from the multi-turn loop rather than retrieval.}
Across the nine model-task pairs outside patient diagnosis, the self-retrieval effect lies within $\pm0.03$ in five cases. For summarization, evidence retrieval, and imaging indication, requiring the agent to retrieve its own evidence changes performance only modestly, with gains no larger than $+0.14$. Once a model is already operating as an agent, giving it a preassembled chart therefore provides little advantage over allowing it to navigate the structured EHR interface itself. By contrast, the agentic-loop effect is negative in all nine of these cases and reaches $-0.237$ for Llama 4 Scout on imaging indication. This penalty becomes larger for weaker models: on imaging indication, the loop effect is only $-0.014$ for GPT~5.3 and $-0.015$ for Mistral Large but $-0.237$ for Llama 4 Scout. Failed sessions also accumulated substantially more context, averaging 145k input tokens compared with 60k for successful sessions, and often ended in malformed outputs. These results suggest that long multi-turn trajectories can exceed the context-management capacity of smaller models, causing performance to degrade even when the required evidence is available.

\textbf{Agents help when the task requires assembling evidence across a longitudinal record.}
Patient diagnosis is the only task for which the self-retrieving agent outperforms the full-context single-turn baseline for all three models. Relative to the full-context LLM, the self-retrieving agent improves severity-weighted F1 by $+0.137$ for GPT~5.3, $+0.141$ for Mistral Large, and $+0.339$ for Llama 4 Scout. The self-retrieving agent also exceeds the full-context agent by 0.10--0.18 for all three models, showing that, on this task, allowing the agent to selectively navigate the longitudinal chart is more effective than supplying the full chart upfront. Patient diagnosis differs from the other tasks because the required evidence is distributed across multiple encounters and must be integrated into a longitudinal problem list. In this setting, tool use helps the model identify and assemble information that is difficult to compress into a single preconstructed prompt.

For summarization, evidence retrieval, and imaging indication, the opposite pattern holds. Their relevant evidence is comparatively localized, and the full-context LLM outperforms the best agentic condition by 0.07--0.15 on summarization and by 0.02--0.19 on retrieval; on imaging indication the gap is within 0.02 for GPT~5.3 and Mistral Large but 0.20 for Llama 4 Scout. The self-retrieval effect is small on these tasks, whereas the multi-turn loop introduces a substantial penalty. This suggests that when the required context can be gathered reliably in advance, a scripted retrieval pipeline followed by single-turn inference may be more effective than a general-purpose agent. Agentic workflows provide the clearest benefit when information gathering and longitudinal assembly are themselves central parts of the task, rather than simply whenever the input contains clinical text.

\section{Extraction and ontology grounding details}
\label{app:grounding}
\renewcommand{\thetable}{\Alph{section}}
\setcounter{table}{0}

This appendix documents stage (2) of the pipeline (Section~\ref{sec:construction}) at the level needed to reproduce it. Constants below are those in the released code; the LLM used throughout is Kimi 2.5 (\texttt{kimi-k2.5}, default sampling, 4{,}096 max output tokens), and every call is keyed by a SHA-256 hash of its input, cached, and written to a call log with token counts and raw output.

\paragraph{Concept extraction.} Each of the 7{,}003 board questions is sent once, with the vignette (truncated to 2{,}000 characters), the correct answer and explanation (500 characters), and the distractors (200 characters each). The prompt asks for a fixed JSON object with four blocks: the \emph{primary diagnosis} (what the correct answer points to), \emph{differential diagnoses} (one per distractor), \emph{secondary diagnoses} (comorbidities, risk factors, and predisposing conditions stated in the vignette but not among the choices), and \emph{clinical findings}. Each diagnosis carries a standard name, an ICD-10-CM suggestion, an organ-system category (21 values), an acuity (acute, chronic, acute-on-chronic, unspecified), and a confidence. Each finding carries a name, one of nine types (symptom, sign, lab value, vital sign, imaging finding, procedure result, history item, medication, demographic), the literal value if any, a presence flag (absent findings are retained as negations), and a relevance label (key, supporting, background, distractor). Outputs that fail structural validation (no primary diagnosis or no finding list) are retried up to three times. The run produced 53{,}746 diagnosis mentions and 138{,}777 finding mentions, which deduplicate to 9{,}623 diagnoses and 36{,}620 findings after the grounding step below. Category, acuity, and finding-type strings are normalized to their controlled vocabularies by a fixed synonym table.

\paragraph{ICD-10-CM resolution.} The LLM's suggested code is treated as a candidate and never accepted on its own. Each diagnosis mention is resolved against the CMS ICD-10-CM FY2025 table (97{,}584 codes, of which 74{,}260 are billable leaf codes) by a deterministic cascade: (i)~the suggested code, after dot normalization, is looked up in the table and accepted if it is a billable code; (ii)~otherwise the diagnosis name is matched case-insensitively against code descriptions; (iii)~otherwise a fuzzy match is attempted, with a token-overlap prefilter over description tokens followed by a \texttt{difflib} sequence-similarity score, accepting the top candidate if its score is at least 0.70; (iv)~otherwise the unvalidated suggestion is retained and flagged; (v)~otherwise the diagnosis is left uncoded. Over the 53{,}746 mentions, 83.2\% resolve by (i), 0.2\% by (ii), 4.3\% by (iii), 7.3\% keep a flagged unvalidated code, and 5.0\% remain uncoded. Mentions are then merged on the resolved code (or on the lower-cased name when uncoded), keeping the highest-priority resolution for each node. Because patient-diagnosis scoring matches at the ICD-10 category level (Section~\ref{sec:construction}), a flagged code that is correct at the three-character level still scores correctly.

\paragraph{SNOMED CT mapping.} SNOMED identifiers proposed by the LLM are discarded. Diagnoses and findings are mapped against the SNOMED CT US Edition (September 2025 snapshot; fully specified names with the semantic tag stripped, plus all active synonyms) in four passes, each applied only to nodes the previous pass left unmapped: (i)~exact description match, choosing among several concepts by sequence similarity to the display name; (ii)~for diagnoses only, the ICD-10-CM code is mapped to SNOMED through the SNOMED-to-ICD-10-CM Extended Map reference set, again picking the most similar concept; (iii)~fuzzy description match at a near-exact threshold (0.95 for diagnoses, 0.90 for findings), after rule-based name normalization for findings (numeric values and units stripped, a trailing ``history'' removed, sex terms canonicalized); (iv)~SapBERT \citep{liu2021sapbert} nearest-neighbour matching, in which the name is embedded and compared by cosine similarity against precomputed embeddings of every active SNOMED concept, accepted at 0.70 for diagnoses and 0.65 for findings. Coverage after all passes is 88.0\% of diagnoses (8{,}469/9{,}623) and 97.5\% of findings (35{,}700/36{,}620).

\paragraph{LOINC mapping.} Findings of type lab value are mapped to LOINC 2.82 without any LLM. The name is expanded into candidate strings by stripping numeric values and units, qualitative prefixes (elevated, decreased, positive, \ldots), trailing qualifiers (level, test, ratio, \ldots), and specimen prefixes (serum, urine, \ldots), and the SNOMED preferred term, if any, is added as a further candidate. Candidates are tried in order against the long common name, then the component, then by fuzzy match at 0.90; when several codes match, the serum/plasma, quantitative, and most established code is preferred. Coverage is 28.9\% (1{,}009/3{,}488); most unmapped lab findings are qualitative composites (e.g., ``anion-gap metabolic acidosis'') that are SNOMED findings rather than LOINC observations and remain grounded through their SNOMED code.

\paragraph{Graph assembly.} Question--diagnosis edges (51{,}075; role correct, distractor, or secondary) and question--finding edges (138{,}777; with value, presence, and relevance) are written directly from the extraction. Diagnosis--finding edges are produced by a second LLM pass: for each of the 2{,}975 correct-answer diagnoses, the findings co-occurring with it across its questions (at most 40 per call) are presented together with their co-occurrence counts, and the model assigns one of six relationship types (pathognomonic, highly suggestive, commonly seen, risk factor, protective, rules out) and an estimated frequency, omitting unrelated findings; this yields 52{,}082 typed edges over 2{,}952 diagnoses. Fact cards are linked by a third LLM pass in batches of ten: the model names the diagnoses and findings each fact describes with a relevance type (defines, differentiates, treatment, epidemiology, mechanism for diagnoses; defines, explains, interpretation, normal variant for findings), and each returned name is resolved to an \emph{existing} graph node by exact match, then substring containment, then fuzzy match at 0.80, so that fact linking never creates new concepts; 46{,}223 of 53{,}999 fact cards receive at least one link (68{,}560 fact--diagnosis and 103{,}879 fact--finding edges). Finally, each vignette is segmented by the LLM into 18 standard EHR section types (demographics, chief complaint, HPI, past medical and surgical history, medications, allergies, family and social history, review of systems, vitals, physical exam, labs, imaging, pathology, other studies, assessment, plan) under an instruction to copy the source text verbatim and add nothing; regular-expression checks for age and sex, vital signs, and laboratory values flag sections the model omitted for review. This produced 51{,}940 sections across 6{,}990 questions. For one source question of the running example, an emergency presentation with polyuria, polydipsia, and confusion, the extraction yields the primary diagnosis hyperosmolar hyperglycemic state (validated to ICD-10-CM E11.01), secondary diagnoses including type 2 diabetes (E11.9) and acute kidney injury (N17.9), and 26 typed findings such as polyuria (symptom, key) and metformin therapy (medication, background). This graph becomes the canonical representation of every patient and serves as the source of all benchmark labels. Appendix Table~\ref{tab:grounding} summarizes node and edge counts and coverage.

\begin{table}[h]
\centering
\small
\begin{tabular}{llrr}
\toprule
\textbf{Layer} & \textbf{Element} & \textbf{Count} & \textbf{Grounded} \\
\midrule
Nodes & Diagnoses & 9{,}623 & ICD-10-CM 75.9\%; SNOMED 88.0\% \\
      & Clinical findings & 36{,}620 & SNOMED 97.5\% \\
      & \quad of which lab values & 3{,}488 & LOINC 28.9\% \\
      & Fact cards & 53{,}999 & 85.6\% linked \\
      & EHR sections & 51{,}940 & 6{,}990 questions \\
\midrule
Edges & Question--diagnosis & 51{,}075 & deterministic \\
      & Question--finding & 138{,}777 & deterministic \\
      & Diagnosis--finding (typed) & 52{,}082 & LLM-typed \\
      & Fact--diagnosis & 68{,}560 & LLM-linked, resolved to existing nodes \\
      & Fact--finding & 103{,}879 & LLM-linked, resolved to existing nodes \\
\bottomrule
\end{tabular}
\caption{Knowledge graph size and ontology coverage after stage (2). ICD-10-CM coverage counts only table-validated codes; a further 14.6\% of diagnoses carry a flagged, unvalidated LLM-proposed code.}
\label{tab:grounding}
\end{table}

\section{Patient clustering and record generation details}
\label{app:generation}

This appendix specifies stage (3) of the pipeline (Section~\ref{sec:construction}) and then traces the running example through it. All constants are those in the released code.

\paragraph{Question nodes.} Every board question becomes a node with: age and sex, parsed by regular expressions from the demographics section produced in stage (2) (falling back to the raw vignette, then to pronouns for sex); an age bucket (pediatric 0--17, young adult 18--35, middle adult 36--60, older adult 61 and over); the set $D(q)$ of its correct and secondary diagnosis identifiers; its organ-system label; an acuity (chronic if any correct diagnosis is chronic or acute-on-chronic, else acute); and smoking and alcohol status parsed from its social-history section, when present. Questions without both an age and a sex cannot be placed in a partition and are not clustered.

\paragraph{Compatibility relations.} Two nodes $a,b$ are \emph{demographically compatible}, $a \sim b$, when all of the following hold: same sex; same age bucket; $|\mathrm{age}(a)-\mathrm{age}(b)| \le \tau$ with $\tau = 2, 5, 7, 10$ years for the four buckets; and no social-history contradiction, defined as one node stating never-smoker and the other current or former smoker, or one stating no alcohol use and the other heavy use. They are \emph{clinically linked}, $a \frown b$, when $D(a)\cap D(b)\neq\emptyset$; distractor diagnoses are excluded from $D$, so a shared wrong answer never links two questions.

\paragraph{Constrained greedy graph clustering.} Within each (sex, bucket) partition, every node's degree is the number of other nodes that are both compatible and linked to it. Nodes are visited in decreasing degree. An unvisited node $s$ seeds a cluster $C=\{s\}$; candidates are the unclustered nodes $c$ with $s \sim c$ and $s \frown c$, ordered by $|D(c)\cap D(s)|$ descending. A candidate is admitted when (i) $c \sim m$ for every $m \in C$ and (ii) $c \frown m$ for at least one $m \in C$. Rule (i) makes $C$ a clique under $\sim$, so a patient never has two encounters with incompatible demographics; rule (ii) requires only that the graph of $\frown$ edges restricted to $C$ be connected, so later encounters may share a diagnosis with an intermediate encounter but not with the seed. Growth stops when $|C|$ reaches a target that depends on the cluster's acuity, re-evaluated after every admission: 3 if all correct diagnoses are acute, 5 if a chronic diagnosis is present within a single clinical organ system, and 8 if chronic diagnoses span several. Clusters of size one are discarded. Patient age is the median of member ages. The procedure is deterministic (no randomness, no LLM) and runs in seconds. On the 7{,}003 questions it produced 1{,}268 patients covering 5{,}602 questions; 1{,}401 questions had no compatible, linked partner or lacked demographics and are not part of the benchmark. Cluster sizes: 350 patients with two encounters, 355 with three, 58 with four, 116 with five, 29 with six, 29 with seven, and 331 with eight.

\paragraph{Profile generation (one LLM call per patient).} The prompt states the patient's age and sex, lists the cluster's correct diagnoses, and for each source question includes only the chief complaint, the first 300 characters of the HPI, and the first 200 characters each of the past medical history, medications, and social history. It asks for a JSON profile with twelve fixed keys (race/ethnicity, occupation, insurance, smoking status, alcohol use, chronic conditions, surgical history, family history, allergies, and home medications with doses) under the constraints that nothing may contradict a source excerpt, that chronic conditions are background comorbidities rather than the diagnoses being tested, and that home medications be appropriate to those conditions. Returned profiles are checked for the required keys and for agreement of age (within five years) and sex with the cluster.

\paragraph{Timeline planning (one LLM call per patient).} Given the profile and, for each encounter, its correct diagnoses with acuity, organ system, chief complaint, and HPI summary, the model orders the encounters and assigns each a visit type (outpatient, emergency, inpatient, ICU, telehealth, procedure, follow-up), a department, a fictional attending, a chief complaint, and a date in months from the first visit, under stated rules (chronic care before acute events, non-decreasing dates, first visit at month 0, every source question used exactly once, span at most ten years). Each response is validated programmatically for these properties; encounters are re-sorted by date if needed, unknown visit types are normalized through an alias table, and dates are anchored to a fixed calendar start. Resulting visit types across the corpus are 3{,}079 outpatient, 1{,}755 emergency, 383 inpatient, 296 ICU, 45 follow-up, 29 procedure, and 15 telehealth.

\paragraph{Note assembly (deterministic) and HPI rewriting (one LLM call per non-first encounter).} Each encounter note is assembled from the stage (2) sections of its source question, in a fixed section order under standard headers, with three deterministic edits: the past medical history is prefixed with an active problem list containing the correct diagnoses of all earlier encounters (with their dates) and the profile's chronic conditions; the profile's home medications are appended to the source medication list when not already present; and allergies, family history, social history, and surgical history are filled from the profile only when the source vignette has no such section. Every section row records its source section identifier and whether it was modified, so provenance is recoverable line by line: of the 59{,}964 released sections, 28{,}433 have no source section (the chief complaints, which come from the timeline step, and sections filled from the profile) and 38{,}550 are marked as modified. For encounters after the first, the source HPI is then rewritten by the LLM under a prompt that supplies the profile, the patient's age advanced by the years elapsed since the first visit, the current encounter's metadata and diagnosis, and a summary of the prior encounters, and that requires the rewrite to preserve every clinical detail of the original, add at most one or two opening sentences of history, introduce no new findings or assessments, and stay within 150\% of the original length; rewrites outside 50--250\% of the original length are flagged. 4{,}211 of the 5{,}602 notes carry a rewritten HPI; the remainder (all first encounters, plus a small number without an HPI section) are purely template-assembled.

\paragraph{Running example.} Three questions, all men in the middle-adult bucket, are linked as follows: question A (a 52-year-old with type 2 diabetes and hypertension presenting with migratory abdominal pain and shock; correct diagnosis perforated appendicitis with septic shock, secondary diagnoses including type 2 diabetes and acute kidney injury), question B (a 58-year-old with type 2 diabetes on metformin and sitagliptin presenting with polyuria and confusion; correct diagnosis hyperosmolar hyperglycemic state), and question C (a 58-year-old with diabetes and chronic kidney disease presenting with fever and hypercapnia; correct diagnosis acute hypercapnic respiratory failure). Ages 52 and 58 fall within the 7-year middle-adult tolerance, all three share the type 2 diabetes and acute kidney injury nodes, and because every correct diagnosis is acute the cluster closes at three encounters. The profile call returned chronic conditions of type 2 diabetes, hypertension, and stage 3b chronic kidney disease, six home medications, a prior appendectomy, a family history of diabetes and renal disease, and former smoking. The timeline call ordered A, B, C as emergency (month 0), emergency (month 8), and ICU (month 16) visits. Note assembly copied each vignette's sections and prefixed encounter B's past medical history with the active problem list ``Acute appendicitis with perforation and septic shock (diagnosed 2020-01-15); Type 2 Diabetes Mellitus; Hypertension; Chronic Kidney Disease Stage 3b''. The HPI rewrite for encounter B opens with the patient's known history and prior septic shock before the source presentation. The benchmark labels never touch this prose: the patient-diagnosis reference is the three correct diagnoses with their acuities and first-encounter dates, the retrieval query for these diagnoses grades the patient's 34 chart sections (13 highly relevant, 11 relevant, 6 marginal, 4 not relevant) from the diagnosis--finding graph, and the imaging item for encounter A pairs its order with a graph-derived clinical question. The patient is in the training split and can be inspected in the released database by its identifier.

\section{Ground-truth construction details}
\label{app:labels}

This appendix specifies stage (4) of the pipeline (Section~\ref{sec:construction}): how each task's labels are computed from the graph, which parts (if any) involve an LLM, and how the running example is labelled. Constants are those in the released code. Throughout, ``the patient's diagnoses'' means the set $D^\ast$ of correct-answer diagnosis nodes of the patient's source questions, and ``the patient's findings'' means the union of the typed findings extracted from those questions, each with its extracted relevance label (key, supporting, background, distractor).

\paragraph{Patient diagnosis (deterministic).} The reference is $D^\ast$ in encounter order. Each entry stores the diagnosis identifier, ICD-10-CM code, SNOMED identifier, display name, acuity, and the identifier and date of the first encounter at which it appears; entries with acuity chronic or acute-on-chronic are listed as chronic conditions and the rest as active diagnoses, and an encounter-to-diagnosis map records which encounter introduced each. Secondary diagnoses (comorbidities stated in the vignettes) are deliberately not in the reference; they are handled at scoring time by the chart-neutral rule of Section~\ref{sec:construction}. Scoring matches predicted and reference codes at the three-character category level and weights each reference entry by a severity tier: 3 (critical) for a fixed list of eleven life-threatening categories (sepsis, myocardial infarction, pulmonary embolism, stroke and intracranial hemorrhage, respiratory failure, acute kidney injury, hepatic failure, fluid and electrolyte disorders, anaphylaxis) or for an acute diagnosis in the infectious, neoplastic, hematologic, circulatory, respiratory, or injury chapters; 2 (moderate) for other acute diagnoses and for chronic diagnoses in those chapters; 1 (routine) otherwise. The primary metric is the harmonic mean of tier-weighted recall and unweighted precision. Difficulty is a fixed score of encounter count, organ-system count, and the co-presence of chronic and acute diagnoses. The reference lists average 3.6 diagnoses.

\paragraph{Evidence retrieval (deterministic).} One item per patient; the query is $D^\ast$ and the corpus is the patient's own encounter sections other than assessment and plan, which may state the answer. Each section is graded from the findings of its encounter's source question that fall in that section (by finding type: vitals to the vitals section, laboratory values to labs, symptoms and history items to the HPI, and so on). For each such finding, let $r$ be its extracted relevance and $e$ the strongest typed edge from the finding to any diagnosis in $D^\ast$. The finding scores 3 if $r$ is key and $e$ is pathognomonic or highly suggestive; 2 if $r$ is key with any other or no edge, or $r$ is supporting and $e$ is commonly seen; 1 if $r$ is supporting or background otherwise; and 0 if it is a distractor or has no relevance label. The section's grade is the maximum over its findings. Across the 1{,}268 patients this yields 58{,}926 graded sections (12{,}759 at grade 3, 29{,}263 at grade 2, 10{,}622 at grade 1, 6{,}282 at grade 0); a section is counted relevant for precision at 5 when its grade is 2 or 3, and NDCG at 10 uses the full 0--3 scale. Fact-card passages linked to $D^\ast$ were graded by the same procedure from their link type but are not part of the released corpus, since retrieval is scored over chart sections only.

\paragraph{Context summarization, unconditioned (deterministic label, LLM narrative).} One item per patient. The scored label is the list of \emph{must-include findings}: every finding extracted with relevance \emph{key} from any of the patient's source questions, deduplicated by name and capped at 20 (mean 18.8). The finding-level score is the fraction of these findings present in the generated summary under an abbreviation- and negation-aware matcher. In addition, the LLM writes a reference narrative from the profile, all encounter notes, and the graph diagnosis list under a prompt asking for a 5--10 sentence chronological synthesis; this narrative is used only for ROUGE-L, a secondary metric, and is never consulted for the finding-level score. The clinical question for all unconditioned items is fixed (``What is the current active problem list and clinical trajectory for this patient?'').

\paragraph{Context summarization, specialty-conditioned (deterministic).} Each diagnosis receives zero or more \emph{home} specialties from a fixed table of ICD-10-CM code ranges to twenty specialties, reviewed by a second clinician (for example, I10--I59 to cardiology, I60--I69 to neurology, G00--G09 to both neurology and infectious disease). For a patient with present diagnoses $D^\ast$, a specialty $S$ is \emph{involved} if some diagnosis in $D^\ast$ is home to $S$, otherwise \emph{absent}; every involved specialty yields an item, and up to two absent specialties yield abstention items whose correct answer is to report no relevant problems. Each key finding of the patient is \emph{owned} by the diagnoses in $D^\ast$ to which it has a typed edge (or, failing that, by the correct diagnosis of its source question), and is tiered for $S$ as: \emph{primary} if an owner is home to $S$; \emph{relevant} if an owner is joined to a present home diagnosis of $S$ by a class-1 definitional SNOMED ``due to'' edge or by one of the physician-curated edges; \emph{neutral} if joined only by a class-2 associative or class-3 shared-finding edge; and \emph{excluded} if joined by no edge at all. Findings owned by a pathognomonic or highly suggestive edge are additionally marked \emph{critical}. The headline score is the harmonic mean of recall over critical primary-and-relevant findings and one minus the leakage rate, where leakage is the fraction of excluded findings the summary mentions; neutral findings are neither credited nor penalized, which keeps the relevance boundary independent of the associative edges whose validity Section~\ref{sec:verifiable} assesses. The corpus holds 6{,}345 such items (3{,}809 involved, 2{,}536 absent).

\paragraph{Imaging indication (LLM-authored from graph-fixed inputs).} Every encounter whose source vignette contains an imaging section is an item (1{,}865 encounters; 675 radiographs, 529 ultrasounds, 342 CT, 179 MRI, 50 CT angiography, 90 other). A single LLM call receives the patient profile, the encounter's non-imaging sections, the imaging section itself, the encounter's correct diagnosis, and the chief complaints of up to five prior encounters, and returns two objects. The \emph{order} (modality, body region, priority, and a 2--8 word indication in clinician shorthand that must not name the diagnosis) is what the evaluated model sees, together with the chart up to that encounter. The \emph{reference} holds the clinical question the ordering clinician is inferred to have had, a 3--5 sentence pre-read summary, the must-include imaging findings, a differential of 2--4 coded diagnoses, and relevant non-imaging data. Responses are validated for the required fields and for indication length. Because the reference question is free text, it is scored by extracting clinical concepts from both prediction and reference with an ontology-grounded matcher and computing concept-level F1 (Section~\ref{sec:results}), so that wording differences between the LLM-authored reference and a model's answer are not penalized. This is the one task whose reference is not a deterministic function of the graph; the LLM is, however, told the correct diagnosis, so the reference question is anchored to the graph label rather than to the model's own reading of the case.

\paragraph{Provenance and exclusions.} Every ground-truth row stores its task, granularity, and the patient, encounter, or question it derives from, together with a difficulty label; every LLM call in this stage is cached by input hash and logged. A post-hoc rule set flags source questions whose ``diagnosis'' is not a clinical entity (biostatistics, study design, ethics) and marks the affected entries as non-diagnostic so that they are skipped at scoring time; in the released v1.3 corpus all 12{,}014 items are diagnostic.

\paragraph{Running example.} For patient 1973, $D^\ast$ is \{perforated appendicitis with septic shock (K35.890, acute), hyperosmolar hyperglycemic state (E11.01, acute), acute hypercapnic respiratory failure (J96.02, acute)\}, so the patient-diagnosis reference has three active and no chronic entries, weighted 2, 2, and 3: K35 and E11 are acute diagnoses outside the critical chapters and therefore moderate, whereas J96 is on the critical list. The retrieval query grades the patient's 34 sections at 13, 11, 6, and 4 for grades 3 to 0; for instance, the labs section of the second encounter is grade 3 because severe hyperglycaemia is a key finding with a highly suggestive edge to hyperosmolar hyperglycemic state. The unconditioned summary must mention 20 key findings, among them abdominal pain, fever, rebound tenderness, leukocytosis, hyperlactatemia, polyuria, polydipsia, altered mental status, severe hyperglycaemia, and ketonuria (the cap of 20 is reached before the third encounter's findings). The specialty variant yields items for the involved specialties endocrinology (primary findings led by severe hyperglycaemia, confusion, and hypotension, all critical), general surgery and gastroenterology (rebound tenderness, leukocytosis, free fluid), and pulmonology (arterial blood gas values, accessory muscle use), plus two absent-specialty abstention items (cardiology, dermatology). The imaging item for the first encounter carries the order ``CT abdomen/pelvis, stat, indication: RLQ pain, fever'' and a reference question asking whether acute or stump appendicitis, diverticulitis, or another perforated viscus is causing septic shock, with a differential of stump appendicitis, diverticulitis, perforated peptic ulcer, and ischemic colitis.

\section{Real and synthetic note examples}
\label{app:noteexamples}

Panels A-C show clinical text used in the realism study (Section~\ref{sec:realism}). Panel~A shows an excerpt from an original MIMIC-IV discharge note, with its original de-identification masks preserved. Because the MIMIC-IV Data Use Agreement restricts redistribution of patient-level clinical text, we reproduce only a shortened, de-identified excerpt here rather than the complete note used in the study. Panel~B shows the same admission after conversion to the Synthetic Hospital note format used during physician review. This normalization preserved the clinical content while removing formatting differences. Panel~C shows the full HPI of a Synthetic Hospital patient with a comparable abdominal presentation. Together, Panels A-B illustrate that records from a different documentation system can be normalized to the Synthetic Hospital format while preserving their clinical content, while Panel~C provides an example of the synthetic documentation evaluated by physicians.

\newtcolorbox{notepanel}[1]{colback=gray!4, colframe=gray!60, title=#1,
  fonttitle=\small\bfseries, fontupper=\footnotesize\ttfamily,
  breakable, left=4pt, right=4pt, top=3pt, bottom=3pt}

\begin{notepanel}{Panel A: Original MIMIC-IV HPI}
\_\_\_ is a \_\_\_ man with history of olignometastatic
adenocarcinoma s/p duodenal and umbilical mass resection and
adjuvant FOLFOX, who is admitted with abdominal pain and concern
for melena.

Patient reports he was walking home from work yesterday morning
when he developed acute epigastric abdominal pain. It comes in
waves up to \_\_\_ aching pain, currently \_\_\_ at baseline. Not
worse with food or activity. Mild nausea last night but none 
now.
Also had black stool this morning, although reports normal
consistency. Otherwise no fevers or chills. No headache,
dysphagia, or odynophagia. No CP or SOB. He has a mild cough x3
weeks. No current nausea. No dysuria. No new rashes or joint
pains. He presented to clinic this morning and he was referred 
to
the ED.
[\textit{ED course --- vitals, laboratory values, and imaging --- omitted}]
\end{notepanel}

\begin{notepanel}{Panel B: MIMIC-IV HPI after format normalization}
[redacted] is a [redacted] man with history of olignometastatic adenocarcinoma s/p duodenal and umbilical mass resection and adjuvant FOLFOX, who is admitted with abdominal pain and concern for melena. Patient reports he was walking home from work yesterday morning when he developed acute epigastric abdominal pain. It comes in waves up to [redacted] aching pain, currently [redacted] at baseline. Not worse with food or activity. Mild nausea last night but none now. Also had black stool this morning, although reports normal consistency. Otherwise no fevers or chills. No headache, dysphagia, or odynophagia. No CP or SOB. He has a mild cough x3 weeks. No current nausea. No dysuria. No new rashes or joint pains. He presented to clinic this morning and he was referred to the ED.
[\textit{ED course --- vitals, laboratory values, and imaging --- omitted}]
\end{notepanel}

\begin{notepanel}{Panel C: Synthetic Hospital HPI}
The patient is a 69-year-old male with a history of neurogenic bladder secondary to sacral nerve root injury initially evaluated in 2023, complicated by 25 years of poorly controlled type 2 diabetes mellitus with peripheral neuropathy and recent diagnosis of gastric adenocarcinoma, who presents to the emergency department with progression of lower urinary tract symptoms. He describes his leakage as continuous dribbling of urine even after voiding, accompanied by a persistent feeling of incomplete bladder emptying. He reports that he frequently finds himself needing to change his underwear multiple times per day and that these symptoms have gradually worsened over time. He denies any burning during urination, hematuria, or suprapubic pain. He also notes that despite making frequent trips to the bathroom, he never feels fully relieved. His wife mentions that he often has difficulty initiating urination and sometimes seems to have a prolonged delay before a strong stream begins. On further questioning, he admits that he previously had episodes suggestive of urinary retention, for which he was intermittently catheterized in the past.
[\textit{ED course --- vitals, laboratory values, and imaging --- omitted}]
\end{notepanel}

\section{Case-mix and distributional validation}
\label{app:casemix}
\renewcommand{\thetable}{\Alph{section}.\arabic{table}}
\setcounter{table}{0}

Synthetic Hospital is designed as an evaluation benchmark rather than a population simulator or training corpus. Its case mix therefore follows the medical-education material from which it is constructed, emphasizing diagnostically informative cases rather than reproducing real-world disease prevalence. To characterize this distribution explicitly, we conducted a pre-registered three-way comparison of ICD-10 chapter distributions across Synthetic Hospital, its source corpus of board-style questions, and a population-calibrated Synthea cohort \citep{walonoski2018synthea} (v4.0.0; 11{,}481 patients). Before analysis, we specified three expectations: Synthetic Hospital should preserve the case mix of its source corpus, differ from the population-oriented Synthea cohort, and retain canonical clinical comorbidity relationships.

\begin{figure}[htbp]
\centering
\includegraphics[width=\textwidth]{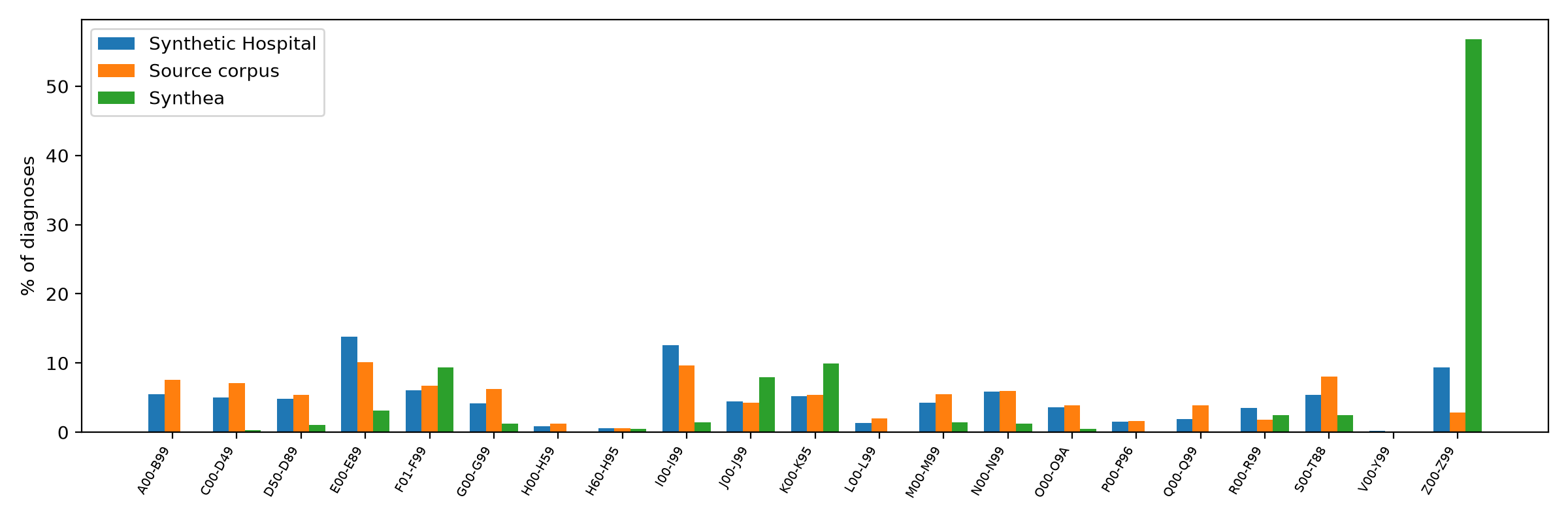}
\caption{ICD-10 chapter distributions across Synthetic Hospital, its source corpus, and Synthea. Synthetic Hospital closely follows the education-derived case mix of its source corpus, whereas the population-oriented Synthea cohort is dominated by Z00--Z99 encounters.}
\label{fig:casemix}
\end{figure}

\textbf{Synthetic Hospital preserves its source case mix but is not population-calibrated.}
All three expectations were supported. Synthetic Hospital closely matches the ICD-10 chapter distribution of its source corpus (Jensen--Shannon divergence [JSD] $=0.029$; Spearman $\rho=0.83$) but differs substantially more from Synthea (JSD $=0.326$; $\rho=0.70$; Table~\ref{tab:casemix}). This difference is systematic rather than random. For example, 56.8\% of Synthea diagnoses fall in the Z00--Z99 chapter covering factors influencing health status and contact with health services, compared with 9.3\% in Synthetic Hospital. Conversely, circulatory and endocrine/metabolic diagnoses are approximately $9\times$ and $4\times$ more frequent, respectively, in Synthetic Hospital (Table~\ref{tab:casemix_chapters}; Figure~\ref{fig:casemix}). These differences reflect the benchmark's intended emphasis on diagnostically informative cases.

\begin{table}[htbp]
\centering
\small
\begin{tabular}{lcc}
\toprule
\textbf{Comparison} & \textbf{JSD} & \textbf{Spearman $\rho$} \\
\midrule
Synthetic Hospital vs.\ source corpus & 0.029 & 0.83 \\
Synthetic Hospital vs.\ Synthea       & 0.326 & 0.70 \\
Source corpus vs.\ Synthea            & 0.436 & --   \\
\bottomrule
\end{tabular}
\caption{Distributional comparison of ICD-10 chapter frequencies across Synthetic Hospital, its source corpus, and Synthea. Synthetic Hospital closely preserves the case mix of its source material while differing substantially from the population-oriented Synthea cohort. Lower JSD indicates more similar distributions.}
\label{tab:casemix}
\end{table}

\textbf{Clinical comorbidity structure is retained in the assembled patients.}
We additionally tested ten canonical comorbidity pairs from the clinical reference set used in Section~\ref{sec:verifiable}. All ten show positive patient-level associations with odds ratios above 4, including T2DM--CKD (14.3), atrial fibrillation--ischemic stroke (16.0), and hypertension--heart failure (9.4) (Table~\ref{tab:casemix_ors}). Thus, the generation process preserves not only marginal disease-category frequencies from its source material but also clinically expected disease co-occurrence in the assembled longitudinal patients.

These analyses characterize the distribution represented by Synthetic Hospital rather than establish population-level fidelity. Agreement with the source corpus is expected because patients are constructed from that corpus; the result verifies that the generation pipeline preserves its intended case mix rather than introducing substantial distributional distortion. Conversely, divergence from Synthea makes explicit that benchmark prevalences should not be interpreted as epidemiologic estimates. This distinction is important for the intended use of Synthetic Hospital: it is designed for zero-/few-shot evaluation of clinical AI systems, not as a synthetic replacement for population-calibrated EHR data for model training or epidemiologic inference.

\textbf{Reproducibility.}
We generated the comparison cohort using Synthea v4.0.0 (jar SHA-256 \texttt{ed43c20a\ldots}) with seed 20260707 and the default Massachusetts configuration, yielding 11{,}481 patients (10{,}000 alive) and 403{,}751 condition records. Because Synthea conditions are SNOMED CT-coded, we mapped them to ICD-10 chapters using the benchmark ontology tables supplemented by a 50-entry manually curated organ-system mapping, achieving 94.1\% coverage of condition rows. The mapping and analysis code are released with the benchmark. The entire comparison is deterministic and uses no language model.

\begin{table}[htbp]
\centering
\small
\begin{tabular}{lrrr}
\toprule
\textbf{ICD-10 chapter} & \textbf{Synth.\ Hospital (\%)} & \textbf{Source corpus (\%)} & \textbf{Synthea (\%)} \\
\midrule
E00-E89 Endocrine/Metabolic & 13.8 & 10.1 & 3.1 \\
I00-I99 Circulatory & 12.6 & 9.6 & 1.4 \\
Z00-Z99 Factors/Health status & 9.3 & 2.9 & 56.8 \\
F01-F99 Mental/Behavioral & 6.0 & 6.7 & 9.4 \\
N00-N99 Genitourinary & 5.9 & 6.0 & 1.3 \\
A00-B99 Infectious & 5.5 & 7.5 & 0.1 \\
S00-T88 Injury/Poisoning & 5.4 & 8.1 & 2.5 \\
K00-K95 Digestive & 5.2 & 5.4 & 9.9 \\
C00-D49 Neoplasms & 5.0 & 7.1 & 0.3 \\
D50-D89 Blood/Immune & 4.8 & 5.4 & 1.1 \\
J00-J99 Respiratory & 4.4 & 4.3 & 7.9 \\
M00-M99 Musculoskeletal & 4.2 & 5.5 & 1.4 \\
G00-G99 Nervous & 4.1 & 6.3 & 1.2 \\
O00-O9A Pregnancy & 3.6 & 3.9 & 0.5 \\
R00-R99 Symptoms/Signs & 3.5 & 1.8 & 2.5 \\
Q00-Q99 Congenital & 1.9 & 3.9 & 0.1 \\
P00-P96 Perinatal & 1.5 & 1.6 & 0.0 \\
L00-L99 Skin & 1.4 & 2.0 & 0.1 \\
H00-H59 Eye & 0.8 & 1.2 & 0.0 \\
H60-H95 Ear & 0.6 & 0.6 & 0.5 \\
V00-Y99 External causes & 0.2 & 0.1 & 0.0 \\
\bottomrule
\end{tabular}
\caption{ICD-10 chapter distributions underlying the case-mix comparison in Table~\ref{tab:casemix}.}
\label{tab:casemix_chapters}
\end{table}

\begin{table}[htbp]
\centering
\small
\begin{tabular}{lr}
\toprule
\textbf{Comorbidity pair} & \textbf{Odds ratio} \\
\midrule
T2DM -- CKD (E11--N18) & 14.3 \\
AFib -- Ischemic stroke (I48--I63) & 16.01 \\
HTN -- Heart failure (I10--I50) & 9.43 \\
Hyperlipidemia -- CAD (E78--I25) & 21.63 \\
COPD -- Respiratory failure (J44--J96) & 70.41 \\
T2DM -- CAD (E11--I25) & 5.16 \\
CKD -- Anemia (N18--D64) & 4.14 \\
Alcoholic liver disease -- Varices (K70--I85) & 100.2 \\
Obesity -- Sleep apnea (E66--G47) & 4.62 \\
HTN -- CKD (I10--N18) & 11.94 \\
\bottomrule
\end{tabular}
\caption{Patient-level associations for ten canonical comorbidity pairs in Synthetic Hospital. Odds ratios use Haldane--Anscombe correction and 3-character ICD-10 categories. All ten pre-specified pairs have odds ratios above 4. Synthea comparisons are omitted because two pairs contain no mapped patients in either category, producing degenerate corrected estimates.}
\label{tab:casemix_ors}
\end{table}

\section{Extended related work}
\label{app:relatedwork}
\renewcommand{\thetable}{\Alph{section}}
\setcounter{table}{0}

This appendix reproduces the full discussion condensed in Section~\ref{sec:related}, together with the synthetic-data paradigm comparison (Appendix Table~\ref{tab:paradigms}).

We situate Synthetic Hospital against four bodies of work: knowledge-QA benchmarks that models saturate but that do not test clinical practice; practice-oriented benchmarks that each cover only part of chart-based work (Table~\ref{tab:coverage}); real-EHR benchmarks that are realistic but gated and ungroundable; and prior synthetic-data paradigms, the closest line of related work (Table~\ref{tab:paradigms}).

\textbf{Knowledge-focused benchmarks and their practice gaps.}
The most-cited medical AI benchmarks are single-vignette multiple-choice or short-answer datasets: MedQA from the United States Medical Licensing Examination (USMLE) \citep{jin2021medqa}, PubMedQA \citep{jin2019pubmedqa}, MedMCQA \citep{pal2022medmcqa}, MultiMedQA/Med-PaLM \citep{singhal2023medpalm}, MMLU-Medical \citep{hendrycks2021mmlu}, BioASQ \citep{tsatsaronis2015bioasq}, HEAD-QA \citep{vilares2019headqa}, MIMIC-derived QA \citep{kweon2024ehrnoteqa,bae2023ehrxqa}, and harder exam-style successors such as MedXpertQA \citep{zhou2025medxpertqa}. That frontier models saturate these formats yet falter on clinical practice is well-established: a review of 39 benchmarks reports 84-90\% accuracy on knowledge versus 45-69\% on practice tasks \citep{gong2025benchmarkreview}, and the multiple-choice format itself inflates competence: models score 64\% on a fictional organ \citep{griot2024glianorex}, drop in free-response form \citep{singh2025freemedqa}, and break under perturbation \citep{cocchieri2026remedqa}, with lifecycle audits and construct-validity analyses concurring \citep{ma2025medcheck,alaa2025constructvalidity}. The open problem is the practice regime these benchmarks cannot reach: the longitudinal, chart-grounded reasoning Synthetic Hospital is built to measure.

\textbf{Practice-oriented benchmarks.}
Recent work moves beyond multiple choice: HealthBench scores rubric-based conversations \citep{arora2025healthbench}, BRIDGE assembles 87 clinical-NLP tasks over real-world clinical text \citep{wu2025bridge}, AgentClinic simulates diagnostic encounters, extended to multimodal tool use in its journal version \citep{schmidgall2024agentclinic,schmidgall2026agentclinic}, MedR-Bench grades multi-stage reasoning \citep{qiu2025medrbench}, and ER-Reason evaluates emergency-room workflows over longitudinal notes \citep{mehandru2025erreason}. Individual chart-based capabilities are also addressed by focused benchmarks: ACI-Bench scores encounter-note generation from simulated dialogues \citep{yim2023acibench}, the ProbSum shared task scores single-admission problem-list summarization from progress notes \citep{gao2023probsum}, EHRSQL evaluates structured querying of EHR databases \citep{lee2023ehrsql}, and clinical summarization with LLMs has been studied with physician reader studies \citep{vanveen2023clinical}. Interactive-diagnosis evaluations (SDBench's sequential NEJM encounters \citep{nori2025sdbench}, AMIE's conversational diagnosis \citep{tu2024amie}) and hospital-scale agent simulacra (Agent Hospital \citep{li2024agenthospital}, AI Hospital \citep{sun2024aihospital}) test complementary interaction skills rather than chart-based work, and MedHELM organizes 35 existing benchmarks into a clinician-validated task taxonomy without adding longitudinal chart tasks \citep{bedi2025medhelm}. Synthetic Hospital is complementary to all of these. Across the five core capabilities of chart-based clinical work (longitudinal reasoning, chart-evidence retrieval, patient problem-list maintenance, patient history summarization and EHR operation \citep{sinsky2016timeallocation,weed1968problemoriented}), several are partially occupied by the benchmarks above, but no prior benchmark combines open data with coverage of all five, and longitudinal problem-list construction as a scored task with its own metric remains, to our knowledge, unoccupied (Table~\ref{tab:coverage}).

\textbf{Real-EHR benchmarks.}
At the other extreme, benchmarks on de-identified real records (MIMIC-III/IV \citep{johnson2016mimiciii,johnson2023mimiciv}, EHRSHOT \citep{wornow2023ehrshot}, TIMER \citep{cui2025timer}, CliniQ \citep{zhao2025cliniq}, the multimodal longitudinal INSPECT cohort \citep{huang2023inspect}, and the broader FHIR-formatted ML landscape \citep{rajkomar2018scalable}) are clinically realistic but reproduce the two barriers we target: they are gated by credentialing and DUAs (PhysioNet for MIMIC, institutional DUAs for EHRSHOT), and their ground truth is only what was charted, so a model failure cannot be separated from an incomplete record. MedAlign is the closest real-EHR instruction benchmark: 983 clinician-written instructions over 276 longitudinal records with clinician reference responses \citep{fleming2024medalign}; it directly exercises longitudinal reasoning and summarization but is DUA-gated, non-redistributable, and its reference responses are subjective clinician text rather than verifiable, ontology-grounded labels. Agent platforms inherit related limits: FHIR-AgentBench \citep{lee2025fhiragentbench} runs on gated MIMIC-IV-FHIR, code-writing agents over MIMIC tables such as EHRAgent \citep{shi2024ehragent} likewise depend on gated data, and MedAgentBench \citep{jiang2025medagentbench}, though openly released with de-identified patient profiles, shares our FHIR framing while evaluating API-level task execution rather than longitudinal, chart-grounded reasoning. Synthetic Hospital provides the same FHIR affordances (1{,}268 patients to MedAgentBench's 100, with the problem-list, summarization, and graded-retrieval tasks they lack) but is open and fully groundable.

\textbf{Concurrent work.} A wave of 2026 benchmarks moves agentic evaluation toward longer-horizon EHR workflows: PhysicianBench instantiates 100 physician-reviewed, long-horizon tasks in a real-record EHR environment \citep{liu2026physicianbench}, EHR-Complex poses 52K interactive SQL/code tasks over MIMIC-IV \citep{qiao2026ehrcomplex}, ClinEnv simulates staged inpatient decision-making over real admissions \citep{lu2026clinenv}, and EHR2Dial-Triage grounds interactive triage conversations in MIMIC-IV-ED \citep{zhu2026elicited}; computer-use variants target clinical and administrative GUIs directly \citep{bedi2026healthadminbench,yu2026medcuabench}. All of these are built on real, access-restricted records or evaluate interface operation rather than chart content, so they are complementary to Synthetic Hospital: none provides openly redistributable longitudinal patient records with constructed, verifiable ground truth, which is the gap this work targets.

\textbf{Synthetic data approaches.}
Generating synthetic records to overcome restricted access to EHR data is well established but existing approaches primarily target privacy rather than benchmark construction. The line begins with GAN-based generation of structured patient records validated by medical-expert review \citep{choi2017medgan}, and extends through neural generation of shareable synthetic clinical notes \citep{melamud2019shareable} and synthetic-note corpora large enough to train openly releasable clinical LLMs \citep{kweon2023asclepius}. Consequently, they do not simultaneously provide open access, complete provenance, clinically realistic narratives, and verifiable ground truth (Table~\ref{tab:paradigms}). Rule-based simulators such as Synthea \citep{walonoski2018synthea} generate standards-compliant FHIR records from predefined disease models but produce structured codes rather than realistic clinical narratives, limiting their use for document-level reasoning tasks. This reflects a difference in objective, not only fidelity: Synthea is designed to simulate population-level health records, whereas Synthetic Hospital is constructed for task-based evaluation with known ground truth; we quantify the resulting case-mix differences in Appendix~\ref{app:casemix}. Statistical and autoregressive generators (EHR-Safe \citep{yoon2023ehrsafe}, HALO \citep{theodorou2023halo}, CEHR-GPT \citep{pang2024cehrgpt}) learn from real EHR distributions. They achieve high fidelity but are trained on protected patient records, retaining the original access restrictions. Moreover, because they reproduce only documented observations, they cannot establish complete ground truth beyond the source documentation. LLM-generated records such as SimSUM \citep{rabaey2024simsum} produce fluent narratives but lack a provenance chain linking statements to underlying clinical facts, making it difficult to distinguish faithful generation from hallucination. Synthetic Hospital instead derives every patient from public educational material rather than real EHRs. Every diagnosis, finding, laboratory result and narrative statement is linked through a typed knowledge graph to ontology-grounded source concepts, yielding complete provenance and benchmark ground truth that does not depend on what happened to be documented in a clinical chart. Closest in spirit, LongHealth \citep{adams2024longhealth} also constructs fictional patients rather than deriving records from real EHRs. However, it comprises 20 single-encounter multiple-choice cases, whereas Synthetic Hospital provides 1,268 longitudinal, ontology-grounded patients with open-ended clinical tasks, complete provenance, and graded ground truth. To our knowledge, Synthetic Hospital is the first benchmark to combine fully synthetic longitudinal EHRs, ontology-grounded provenance, verifiable ground truth and unrestricted open access.

\begin{table}[H]
\centering
\small
\newcolumntype{B}{>{\raggedright\arraybackslash\hsize=1.4\hsize}X}   
\newcolumntype{P}{>{\raggedright\arraybackslash\hsize=0.6\hsize}X}  
\setlength{\tabcolsep}{3pt} 
\begin{tabularx}{\textwidth}{@{} B P c c c c @{}}
\toprule
\textbf{Benchmark} & \textbf{Paradigm} & \textbf{No Real Data} & \textbf{Provenance} & \textbf{Ground truth} & \textbf{Narr.} \\
\midrule
Synthea \citep{walonoski2018synthea} & Rule-based & Yes & Module logic & API responses & No \\
\addlinespace
EHR-Safe \citep{yoon2023ehrsafe} & GAN / statistical & No & None & Statistical & No \\
\addlinespace
\makecell[l]{HALO \citep{theodorou2023halo} \\ CEHR-GPT \citep{pang2024cehrgpt}} & Autoregressive & No & None & Distributional & No \\
\addlinespace
SimSUM \citep{rabaey2024simsum} & LLM-generated & Yes & Partial & Annotations & Yes \\
\addlinespace
Zhou et al.\ \citep{zhou2026clinical} & Knowledge-grounded & No & Partial & Statistical & No \\
\midrule
\textbf{Synthetic Hospital (ours)} & \textbf{Knowledge graph} & \textbf{Yes} & \textbf{Full} & \textbf{Ontology-grounded} & \textbf{Yes} \\
\bottomrule
\end{tabularx}
\caption{Synthetic clinical-data generation paradigms. All narrative approaches (SimSUM, ours) use an LLM to produce prose; the distinction is provenance of the ground truth. SimSUM annotates its own generated text, so generation errors can enter the labels; our ground truth is derived from the ontology-grounded graph independently of the generated narrative, so narrative errors cannot corrupt it. ``No Real Data'' marks approaches that can be built without any access to real patient records (Yes) or that require them (No). The Synthetic Hospital is the only approach that combines full provenance, precise ground truth, rich clinical narratives and fully synthetic data. 
}
\label{tab:paradigms}
\end{table}

\end{document}